\documentclass[a4paper,11pt]{article}

\usepackage{mathpazo,bm}    
\usepackage{amsmath}
\usepackage{cite}
\usepackage[all]{xy}
\usepackage{rotate}
\usepackage{multirow}
\usepackage{lscape}
\usepackage[latin1]{inputenc}
\usepackage{wasysym}
\usepackage{colortbl}
\usepackage{url}
\usepackage[pdftex]{graphicx}
\usepackage{rotating}
\usepackage{makeidx}  
\usepackage{amsfonts}
\usepackage{color}
\usepackage{wrapfig}
\usepackage{algorithmic}
\usepackage{algorithm}

\title{Nonlinearities and Adaptation of Color Vision from Sequential Principal Curves Analysis}

\author{Valero Laparra, Sandra Jim\'enez, Gustavo Camps-Valls and Jes\'us Malo\\
Image Processing Laboratory (IPL), Universitat de Val\`encia.\\
Catedr\'atico A. Escardino -- 46980 Paterna, Val\`encia, Spain.\\
{\tt valape, sjimenez, gcamps, jmalo}@uv.es, \\
{\tt http://isp.uv.es}
\thanks{Published: Neural Comput. 2012 Oct; 24(10):2751-88. DOI: 10.1162/NECO\_a\_00342}
\thanks{Image Processing Laboratory (IPL), Universitat de Val\`encia, Catedr\'atico A. Escardino - 46980 Paterna, Val\`encia (Spain). E-mail: \{valero.laparra, jesus.malo, gcamps\}@uv.es, http://isp.uv.es}
\date{}
}

\begin{document}

\pagestyle{myheadings}
\markboth{Neural Comput. 2012 Oct;24(10):2751-88. doi: 10.1162/NECO_a_00342}{Laparra et al., 2012}
\maketitle


\begin{abstract}
Mechanisms of human color vision are characterized by two phenomenological aspects: the system is nonlinear and adaptive to changing environments.
Conventional attempts to derive these features from statistics use \emph{separate} arguments for each aspect.
The few statistical approaches that do consider both phenomena simultaneously follow {\em parametric formulations} based on empirical models.
Therefore, it may be argued that the behavior does not come directly from the color statistics but from the convenient functional form adopted.
In addition, many times the whole statistical analysis is based on {\em simplified databases} that disregard relevant physical effects in the input signal, as for instance by assuming flat Lambertian surfaces.

In this work, we address the \emph{simultaneous} statistical explanation of (i) the nonlinear behavior of achromatic and chromatic mechanisms in a fixed adaptation state, and (ii) the change of such behavior, i.e. adaptation, under the change of observation conditions. Both phenomena emerge directly from the samples through a single {\em data-driven} method: the Sequential Principal Curves Analysis (SPCA) with local metric. SPCA is a new manifold learning technique to derive a set of sensors adapted to the manifold using different optimality criteria. Moreover, in order to reproduce the empirical adaptation reported under D65 and A illuminations, a \emph{new database} of colorimetrically calibrated images of natural objects under these illuminants was gathered, thus overcoming the limitations of available databases.

The results obtained by applying SPCA show that the psychophysical behavior on color discrimination thresholds, discount of the illuminant and corresponding pairs in asymmetric color matching, emerge directly from realistic data regularities assuming no \emph{a priori} functional form.
These results provide stronger evidence for the hypothesis of a statistically driven organization of color sensors.
Moreover, the obtained results suggest that color perception at this low abstraction level may be guided by
an error minimization strategy rather than by the information maximization principle. \\

\end{abstract}


\newpage

\section{Introduction}

Human color vision is mediated by achromatic and opponent chromatic mechanisms with two fundamental properties~\cite{Fairchild05,Brainard05,Abrams07}: (i) their response is nonlinear given some fixed observation conditions (e.g. illuminant, spatial context, etc.); and (ii) they are able to compensate changes in the observation conditions to keep the perception (color) of the objects constant despite the changes in the input. On the one hand, the nonlinear response of the mechanisms is revealed by the non-uniform nature of discrimination thresholds throughout the tristimulus space~\cite{Stiles82,Cole90,Gegenfurtner92,Romero93}. On the other hand, the adaptation ability of these mechanisms is revealed by asymmetric color matching experiments~\cite{Breneman87,Webster91,Luo91,Luo99}: different physical inputs give rise to the same perception (corresponding stimuli) for equivalent shifts in the context.

The standard empirical models of color vision assume that the system is formed by three linear photoreceptors sensitive to long, medium and short wavelengths (LMS). These models try to reproduce the above mentioned effects using three basic ingredients as reported in~\cite{Fairchild05,Brainard05,Abrams07}: (i) context dependent weighting of the sensitivity of LMS mechanisms, also known as Von Kries normalization, (ii) linear transform to an opponent color space, and (iii) nonlinear saturation of the achromatic and the opponent-chromatic responses.

Following the classical suggestion by Barlow on the relation between image statistics and neural behavior \cite{Barlow61,Barlow01}, a large body of literature argues that mechanisms underlying the perception of object colors are organized according to the statistical regularities of the signals confronted by the sensory systems. However, very often, the proposed statistical approaches deal with color discrimination and color constancy in a {\em separated way}: they are not able to address both adaptation and nonlinearities jointly.

On the one hand, linear approaches based on decorrelation (linear principal component analysis, PCA) and higher order redundancy removal (linear independent component analysis, ICA) explain the existence of spectrally opponent chromatic channels with the right spatial sensitivity. The seminal work of Buchsbaum and Gottschalk~\cite{Buchsbaum83} derives opponent channels from PCA applied to LMS signals subject to a rough model of natural radiances (white noise). Atick et al.~\cite{Atick92} derive the achromatic and chromatic Contrast Sensitivity Functions using decorrelation arguments constrained with error minimization and an idealized model of spatio-spectral radiances. Ruderman et al. \cite{Ruderman98} obtain Fourier-like spatio-chromatic opponent sensors using PCA on the LMS signals obtained from real reflectance measurements. Wachtler et al.~\cite{Sejnowski01} apply linear ICA pixel-wise and patchwise on real hyperspectral photographic images and obtain better coding results than using PCA. Doi et al.~\cite{Doi03} use PCA and ICA to derive spatio-chromatic properties of the lateral geniculate nucleus (LGN) and the primary visual cortex (V1).
Even though the above techniques do not explicitly address adaptation, Webster and Mollon~\cite{Webster97} show that the mean shift (or chromatic adaptation), and the covariance shift (or contrast adaptation) can be roughly reproduced by dimension-wise normalization of the LMS responses, and PCA followed by whitening using the set of adaptation colors under different illuminants. Their work combines an extension of the measurements reported in~\cite{Webster91} (which used single-color adaptation) with the decorrelation-oriented explanation given by Atick et al.~\cite{Atick93}. The obvious problem with linear methods is that they cannot explain non-uniform discrimination, i.e. the nonlinear response.

On the other hand, color discrimination and the associated nonlinearities have been statistically addressed from a different point of view. In this case, the key is the consideration of the limited resolution of any physical sensor and its optimal design to deal with non-uniformly distributed signals. Two different criteria have been proposed in this context. First, Laughlin \cite{Laughlin83} argued that limited resolution mechanisms should be designed to maximize the information transfer  ({\em infomax} approach). In noise-free scenarios, the {\em infomax} principle leads to component independence and nonlinear responses related to the marginal probability density functions (PDFs) \cite{Bell95}. Second, MacLeod et al. proposed that, in order to minimize the representation error in the presence of neural noise (\emph{error minimization} approach), the response of the color sensors should be related to some power of the marginal PDFs in the color opponent directions~\cite{Twer01,MacLeod03,MacLeod03b}. The same reasoning has been applied to explain physiological nonlinearities at the LGN~\cite{Goda09}.
Unfortunately, in these studies, an explicit multidimensional data-driven algorithm to get the optimal set of sensors remained unaddressed: in their experiments they just showed marginal PDFs in predefined linear axes~\cite{Twer01,MacLeod03,MacLeod03b,Goda09}.
And more importantly, no attempt has been made to explain adaptation using {\em infomax} or {\em error minimization}. Again, just one of the aspects, the nonlinear behavior, was addressed by these techniques.

Among the statistically inspired approaches, a remarkable exception to the separate consideration of color discrimination and adaptation is the work by Abrams et al.~\cite{Abrams07}, where the authors investigate whether optimizing the nonlinearity is compatible with optimal color constancy or adaptation. In that case the presented model already has the appropriate {\em parametric formulation} inherited from the empirical models~\cite{Brainard05}. Therefore, it may be argued that, even though the model was statistically fitted, the expected behavior is somehow imposed in advance through the use of a convenient functional form. According to this, it is not shown that the empirical behavior emerges directly from data, but the (also interesting) fact that the empirical models may be statistically optimal in color discrimination and adaptation at the same time.

Finally, we should that many of the above statistical studies rely on {\em simplified databases}. In particular, most of the works dealing with adaptation to changes in the illuminant usually assume that the input radiance is just the product of the spectral reflectance and the illuminant radiance: they assume flat Lambertian surfaces. Therefore, relevant nonlinear phenomena in the image formation process, such as mutual illumination in rough surfaces, are neglected.

In this work, we address the {\em simultaneous} explanation of (i) the nonlinear behavior of achromatic and chromatic mechanisms in a fixed adaptation state; and (ii) the change of such behavior, i.e. adaptation, under the change of observation conditions. This is done by proposing a single non-parametric method based on Principal Curves (PCs) \cite{Hastie89,Delicado01,Einbeck05}: the Sequential Principal Curves Analysis (SPCA). The method exploits the flexibility of PCs to find adaptive (eventually non-linear) sensors.
Moreover, SPCA is equipped with tunable local metric so that the proposed analysis may follow either the {\em infomax} or the {\em error minimization} principles. Given the fact that psychophysical adaptation data are given under D65 and A illuminations \cite{Luo91,Luo99}, in this work, the statistical analysis is made on a {\em new database} consisting of colorimetrically calibrated images of natural objects under these calibrated illuminations. SPCA reproduces the psychophysical behavior on color discrimination thresholds, discount of the illuminant and corresponding pairs in asymmetric color matching. These color vision properties are demonstrated to emerge directly from realistic data regularities without assuming any {\em a priori} functional form. Moreover, the results suggest that color perception at this low abstraction level may follow an error minimization strategy, as suggested by MacLeod~\cite{Twer01,MacLeod03,MacLeod03b}, instead of the information maximization principle suggested in Laughlin~\cite{Laughlin83}.

The remainder of the paper is organized as follows. Section \ref{motivation} motivates the statistical study of color vision by reviewing (i) the basic features of color PDFs and their changes, and (ii) the non-linear and adaptive nature of color vision mechanisms. Section \ref{Learning} reviews the results on {\em infomax} and {\em error minimization} in {\em unsupervised manifold learning}.
This motivates our computational approach, which is presented in Section \ref{generalSPCA}: the Sequential Principal Curves Analysis (SPCA) with local metric.  Section \ref{setup} describes the design of the experiments: the database of calibrated natural color images used and how the simulation of color vision phenomena from SPCA is carried out. Section \ref{nonlinear_results} shows how the proposed statistical technique simultaneously reproduces experimental data on color discrimination and adaptation, compares its performance with empirical color appearance models (CIELab, LLab, CIECAM), and discusses the biological implications of the results. Finally, Section \ref{conclusions} presents the conclusions of the work.


\section{Facts on Color PDFs and Color Mechanisms Behavior}
\label{motivation}

This section motivates the problem of characterizing nonlinearities and adaptation of the color vision mechanisms. First, we review the special characteristics of the color manifolds. Then, the considered perception phenomena are reviewed through two sets of results: (i) the nonlinear behavior of achromatic and opponent chromatic mechanisms~\cite{Stiles82,Cole90,Gegenfurtner92,Romero93}; and (ii) the ability to compensate changes in spectral illumination according to experiments on corresponding colors~\cite{Breneman87,Luo91,Luo99}.

\subsection{Non-uniformities and shifts in color manifolds}

\begin{figure}[h!]
\begin{center}
\setlength{\tabcolsep}{3pt}
\begin{tabular}{|ccccc|}
\hline
D65 & A & D65 zoom  & A zoom & PCA+whitening \\
\includegraphics[height=3cm,width=3cm]{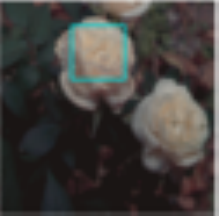} &
\includegraphics[height=3cm,width=3cm]{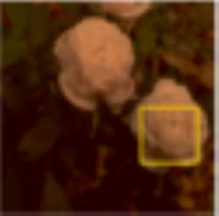} &
\includegraphics[height=3cm,width=3cm]{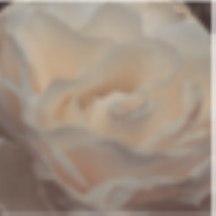} &
\includegraphics[height=3cm,width=3cm]{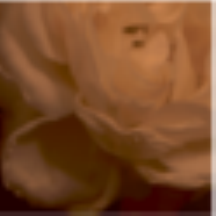} &
\includegraphics[height=3cm,width=3cm]{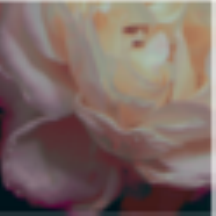} \\
\multicolumn{3}{|c}{LMS distribution} & \multicolumn{2}{c|}{CIExy} \\
\multicolumn{3}{|c}{\includegraphics[height=4.9cm,width=7.60cm]{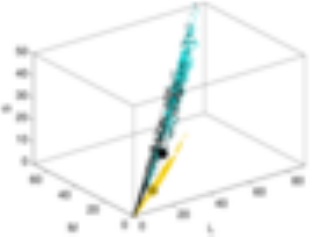}} &
\multicolumn{2}{c|}{\includegraphics[height=4.9cm,width=4.9cm]{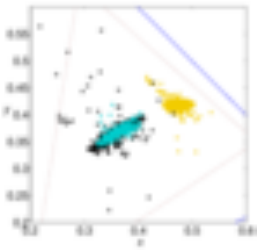}}\\
\hline
D65 (11$^o$) & A (22$^o$) & D65 zoom & A zoom & LS approximation \\
\includegraphics[height=3cm,width=3cm]{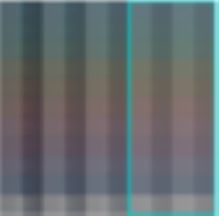} &
\includegraphics[height=3cm,width=3cm]{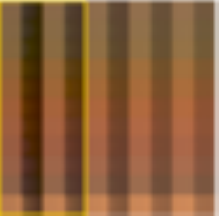} &
\includegraphics[height=3cm,width=3cm]{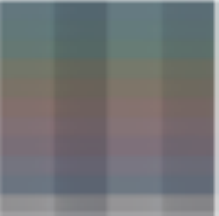} &
\includegraphics[height=3cm,width=3cm]{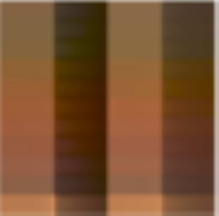} &
\includegraphics[height=3cm,width=3cm]{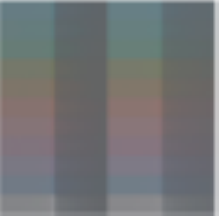} \\
\multicolumn{3}{|c}{LMS distribution} & \multicolumn{2}{c|}{CIExy} \\
\multicolumn{3}{|c}{\includegraphics[height=4.9cm,width=7.59cm]{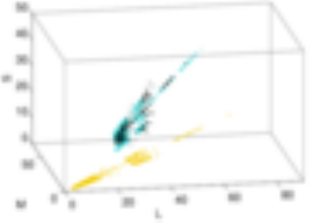}} &
\multicolumn{2}{c|}{\includegraphics[height=4.9cm,width=4.9cm]{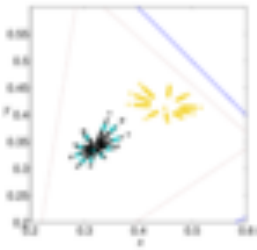}} \\
\hline
\end{tabular}
\end{center}
\vspace{-0.45cm}\caption{\small Illustration of nonlinearities and shifts in color manifolds. Images, tristimulus distributions in LMS and CIExy chromatic coordinates are shown for real (top panel) and synthetic (bottom panel) examples under different spectral illuminations and geometry. In the CIExy diagrams, the triangle of available colors in a standard display is plotted in red for convenient reference. {\bf Top:} Flower images under diffuse D65 (blue dots) and A (yellow dots) illuminant. The synthesized colors (black dots and right image) are obtained from linear color manifold matching by PCA plus whitening. A particular color (large yellow dot) and its linearly obtained corresponding pair (large black dot) have been highlighted for convenient comparison with equivalent illuminant compensation results that will be shown in Section \ref{procedure}. {\bf Bottom:} Lambertian undulated surface under D65 and A illuminants tilted 11$^o$ and 22$^o$, respectively. Here, the synthesized image is obtained by fitting the best least squares linear regression (black dots). }
\label{real_example}
\end{figure}

Observation of the natural world gives rise to measurements that typically live in low dimensional manifolds. The shape of these manifolds and its PDF depend on the interesting features of the underlying phenomenon. For example, in vision, it is known that the spectral reflectance of objects is intrinsically low dimensional~\cite{Maloney86}. Moreover, the physical constraints on the reflectance as well as the geometry of the surfaces give rise to particular statistics of the tristimulus values~\cite{Adelson07,Koenderink10}. However, these distributions are also modified by additional, eventually non-interesting, causes~\cite{Funt93}: both mutual illumination in surfaces of complex geometry and changes in the spectral illumination and its  geometry introduce nonlinear changes in the PDF of the tristimulus values that are difficult to characterize. The examples in Fig.~\ref{real_example} show real (top) and synthetic (bottom) examples of the nonlinear nature of changes in color distributions, which clearly cannot be compensated by linear transforms.

Top panel in Figure~\ref{real_example} illustrates the basic features of the tristimulus PDF in natural scenes and its changes:
\begin{itemize}
\item The distribution is remarkably non-uniform, i.e. densely populated around the color direction determined by the illuminant, and of decreasing density for higher saturation values (see LMS distribution).
\item It displays a strong correlation between the LMS values.
\item Different illumination geometry gives rise to different data distribution along the principal axes. In particular, the highlighted regions differ in the illumination angle: the scene illuminated with D65 (A) is relatively more populated in the high (low) luminance region.
\end{itemize}

The change in orientation of the PDF due to the change in the spectral radiance of the illuminant can be approximately compensated by a rotation, and the change in total radiance by a suitable scaling~\cite{Webster97,Atick93}. However, the classical {\em PCA plus whitening} linear transform is not able to compensate the uneven data distribution along the principal axes. In the example, black dots represent the prediction of the data under D65 from A data using PCA+whitening. In this case, the transformed data are still relatively more concentrated in the low luminance region. Nonlinear transforms either in D65 or A data would be required to equalize the distributions along the individual subspaces. The effect of such equalization would remove the shaded regions that remain in the linearly compensated image.

The inability of linear transforms can be better stressed in a synthetic example with controlled reflectance, surface and illumination geometries.
The bottom panel in Fig.~\ref{real_example} shows a triangularly undulated surface illuminated with D65 and A radiances from different angles (11$^o$ and 22$^o$). In this synthetic case, a one-to-one correspondence between the colors can be established so the linear transform can be optimized to minimize the least squares error, which is not possible in real examples. The result of such optimal linear transform (not restricted to be based on rotations) is represented by the black dots and by the reconstructed image. In this case, the best possible linear transform is unable to compensate for the color changes induced by the change in spectral radiance, surface and illumination geometries: it can shift the yellowish colors to the region of the gray-blueish colors but, as a byproduct, the low luminance colors are markedly desaturated. In this example, the nonlinear change in the PDF
comes from the differences in the surface and illumination geometries.
This stresses the fact that general changes occurring in tristimulus values of natural surfaces are nonlinear.

\subsection{Nonlinear behavior of achromatic and opponent chromatic mechanisms}

The sensitivity of some underlying sensory mechanism is related to its discrimination ability according to the classical Fechner's hypothesis used in psychophysics~\cite{Laming97,Brainard05}: given a unidimensional measure $x$ and an eventually nonlinear response $R$, the bigger the incremental threshold (or just noticeable difference), $\Delta x(x)$, the smaller the slope (or sensitivity) of the underlying mechanism,
\begin{equation}
      \frac{dR(x)}{dx} \propto \frac{1}{\Delta x(x)}.
\end{equation}
Therefore, the response can be estimated from the experimental incremental thresholds by
\begin{equation}
      R(x) = R(x^o) + \beta \int_{x^o}^{x} \frac{1}{\Delta x(x')} dx',
      \label{Fechner}
\end{equation}
where $\beta$ is an irrelevant scaling factor. As a result, if the incremental thresholds are not constant over the stimulus range, the response of the underlying mechanism is nonlinear.

\begin{figure}[t!]
\vspace{-1cm}
\begin{center}
\begin{tabular}{ccc}
\includegraphics[height=3.3cm,width=4cm]{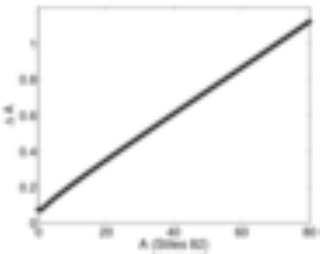} &
\includegraphics[height=3.3cm,width=4cm]{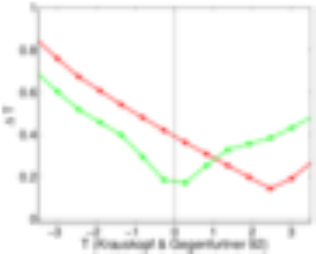} &
\includegraphics[height=3.3cm,width=4cm]{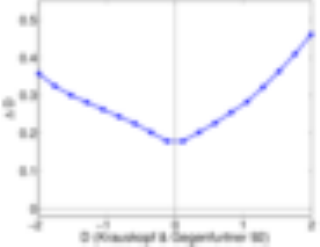} \\
\includegraphics[height=3.3cm,width=4cm]{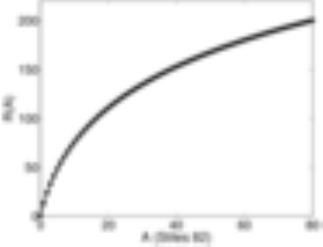} &
\includegraphics[height=3.3cm,width=4cm]{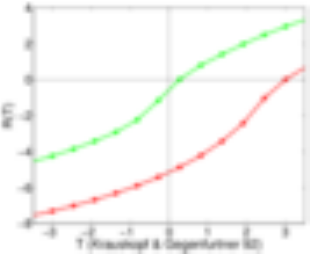} &
\includegraphics[height=3.3cm,width=4cm]{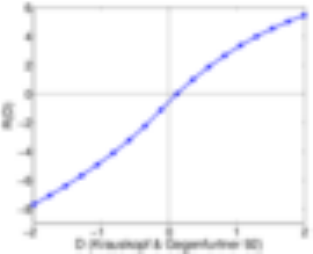} \\
\end{tabular}
\begin{tabular}{cc}
\includegraphics[height=4.2cm,width=4.2cm]{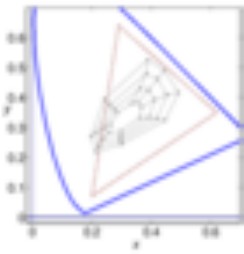} &
\includegraphics[height=4.2cm,width=4.2cm]{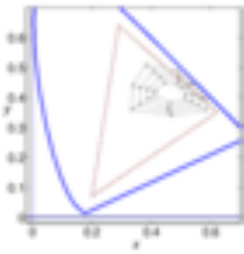}\\
\end{tabular}
\end{center}
\vspace{-0.6cm}\caption{\small Summary of psychophysical results. Experimental color discrimination thresholds in ATD channels (top row) and the corresponding nonlinear responses (middle row). From left to right: channel A (Achromatic), channel T (Green-Red), and channel D (Yellow-Blue). In the case of the T channel, the green and the red curves represent the thresholds in two adaptation conditions: white adaptation point and reddish adaptation point respectively. Bottom row shows the experimental corresponding colors (CIE xy chromaticities) under CIE D65 illuminant (left) and under CIE A illuminant (right).}
\label{non_linearities}
\end{figure}

Figure \ref{non_linearities} (top panel curves) shows the experimental behavior of the achromatic (A), red-green (T), and yellow blue (D) mechanisms.
Top left plot shows the experimental increase of the luminance thresholds (data derived from Fig. 7.10.1 in~\cite{Stiles82}, page 569). Luminance has been expressed in terms of the Ingling and Tsou color space for appropriate comparison to the theoretical predictions below. The increase in thresholds is classically known as the Weber's law~\cite{Stiles82}. Middle left figure shows the saturating nonlinearity of the underlying brightness perception mechanism using Eq.~\eqref{Fechner}. Top center and top right plots show the V-shaped curves of the color incremental thresholds for red-green and yellow-blue stimuli (replotted from Figs. 10 and 11 in~\cite{Gegenfurtner92}). Axes in the figures have been expressed in absolute T and D units by re-scaling the original data (given in threshold relative units) using threshold values for appropriate comparison to the theoretical predictions. Krauskopf and Gegenfurtner~\cite{Gegenfurtner92} used two adaptation conditions: (i) white adaptation point (at the origin in the T and D axes); and, in the case of the T discrimination, they also used (ii) a reddish adaptation point (about 2.5 in the rescaled T axis). Middle central and right plots show the corresponding response curves using Eq.~\eqref{Fechner}.

The data show two interesting features of T and D color perception mechanisms:
\begin{itemize}
\item The discrimination is optimal (minimum threshold) at the chromaticity of the adaptation point, and then the sensitivity decreases as one departs from it. This gives rise to the saturating sigmoidal response curves with maximum slope at the adaptation point.
\item A change in the adaptation state implies a shift in the response curve in order to set the maximum sensitivity region at
the new adaptation point.
\end{itemize}

\subsection{Adaptation and corresponding pairs}
\label{adapt_results}

Corresponding colors are described by two sets of tristimulus values that give rise to the same perceived color when the two samples are viewed in different environments~\cite{Breneman87,Luo91,Luo99}. Corresponding colors reveal the color constancy ability of human observers under change of illumination conditions: despite the change in the linear measurements, the corresponding pairs are perceived as equal. A chromatic adaptation transform should be able to predict corresponding colors.
The chromatic diagrams in Fig.~\ref{non_linearities} show the data compiled by Luo et al.~\cite{Luo91,Luo99} regarding corresponding colors under CIE D65 and CIE A illuminants\footnote{Data available on-line at \texttt{http://colour.derby.ac.uk}}. Note that the corresponding stimuli under CIE A illuminant have reddish-yellowish chromaticities with regard to those under CIE D65 indicating the illuminant compensation ability of human viewers.


\section{Sensor Design by Learning Nonlinear Data Representations}
\label{Learning}

The responses of sensory systems devoted to describe some phenomenon of interest have to convey as much information as possible about the phenomenon while minimizing the representation error for every possible input. The first of these related, but not exactly equivalent, requirements is usually known as {\em Information Maximization}~\cite{Laughlin83,Linsker88,Bell95,Lee00}. 
Here we will refer to the second requirement~\cite{Lloyd82,Twer01,MacLeod03,MacLeod03b} as {\em Error Minimization}.

Additionally, the sensors should discount variations in linear measurements coming from non-interesting sources: even though eventual modifications of the measurement conditions may give rise to changes in the PDF of the linear measurements, such as the ones illustrated in Fig.~\ref{real_example}, the internal representation (the perception) has to be invariant to these changes. In the psychophysics literature, this ability is usually known as {\em adaptation}~\cite{Fairchild05}, in the signal processing literature as {\em adaptive modeling and filtering} \cite{Haykin02}, while in the machine learning literature this has been recently referred to as {\em domain adaptation}~\cite{Storkey09,Pan09}.

The manifold learning method proposed in this paper is motivated by the {\em infomax} and the {\em error minimization} principles, and by the need of learning systems capable of dealing with the {\em adaptation} or {\em dataset shift problem} in the specific context of the color statistics.

\subsection{Nonlinear sensory systems design: infomax and error minimization principles}
\label{sect_infomax}

Processing input observations ${\bf x} \in \mathbb{R}^d$ requires the design of an {\em appropriate} set of $d$, sensors that respond according to the mapping $R$, which  transforms points ${\bf x}$ to ${\bf r} \in \mathbb{R}^d$. Physical sensors may have limited resolution or may be subject to internal noise in such a way that the responses are corrupted according to a sort of quantization $Q$, 
\begin{equation}
\xymatrix{
{\bf x} \ar@/^/[r]^{R} \ar@/_1pc/@{<-}[rr]_{R^{-1}}
 & {\bf r} \ar@/^/[r]^{Q} & {\bf r^\star}}.
\label{generic_transform}
\end{equation}

The {\em infomax} sensory organization principle states that the mapping $R$ has to be selected to maximize the transferred information from ${\bf x}$ to ${\bf r^\star}$. This requirement induces different constraints on the Jacobian of the response transform $\nabla R({\bf x})$ in the noise-free and noisy scenarios depending on the PDF of the input measurements, $p({\bf x})$, as will be reviewed below and extensively reported elsewhere~\cite{Laughlin83,Bell95,Lee00,Lloyd82,Gersho92,MacLeod03}. In general, the sensitivity of the system (the slope of the response) in each region of the input space has to be related to the population in that region. Additionally, assuming that the sensory system considers the internal representation to be Euclidean, as done in psychophysics~\cite{Laming97,Brainard05}, the system induces a {\em perceptual} metric in the input domain, $M({\bf x})$, related to the Jacobian of the response transform, see~\cite{Dubrovin82,Epifanio03,Malo06a,Laparra10a}:
\begin{equation}
M({\bf x}) = \nabla R({\bf x})^\top \cdot \nabla R({\bf x}),
\label{perceptual_metric}
\end{equation}
which follows from considering an Euclidean metric in the response domain, i.e. the sensory system considers all distortions in the same way\footnote{
In the situation described in Eq.~\eqref{generic_transform}, distances induced by small distortions in the input and the response domains, $\Delta {\bf x}$ and $\Delta {\bf r}$, may be described by local metrics $M({\bf x})$ and  $M({\bf r})$:
$d({\bf x}, {\bf x}+\Delta {\bf x})^2 = \Delta {\bf x}^\top \cdot M({\bf x}) \cdot \Delta {\bf x} = \Delta {\bf r}^\top \cdot M({\bf r}) \cdot \Delta {\bf r} = d({\bf r}, {\bf r}+\Delta {\bf r})^2$. Assuming that the response transform $R$ is differentiable, the distortion in the response may be approximated by $\Delta {\bf r} \approx \nabla R({\bf x}) \cdot \Delta {\bf x}$, which, assuming $M(\mathbf{r}) = I$, leads to Eq.~\eqref{perceptual_metric}.}.
Accordingly, relations between the sensitivity of the system and the population of the input space will give rise to relations between the induced metric in the input space and $p({\bf x})$.
The information maximization and error minimization criteria impose different restrictions on both the Jacobian and the metric.

On the one hand, in the noise-free case, the {\em infomax} principle to set $R$ reduces to looking for transforms that lead to responses with maximal entropy or equivalently, to independent responses~\cite{Bell95,Lee00}. This scenario implies a restriction on the Jacobian of the transform:
\begin{equation}
|\nabla R({\bf x})| \propto p({\bf x}),
\label{noise_free_jacobian}
\end{equation}
which, according to Eq.~\eqref{perceptual_metric}, leads to the following determinant of the induced metric:
\begin{equation}
|M({\bf x})| \propto p({\bf x})^2.
\label{noise_free_metric}
\end{equation}
On the other hand, the minimization of the representation error in sensory systems subject to internal noise or limited resolution leads to a different constraint on the Jacobian. In particular, in~\cite{MacLeod03}, MacLeod and Twer show that, in that situation, the optimal sensitivity in mean-square-error terms has to be,

\begin{equation}
|\nabla R({\bf x})| \propto p({\bf x})^{1/3},
\label{noisy_jacobian}
\end{equation}
which is consistent with the classical optimal MSE distribution of discrete perceptions in Vector Quantization~\cite{Lloyd82,Gersho92}. According to Eq.~\eqref{perceptual_metric}, the determinant of the induced metric should be:
\begin{equation}
|M({\bf x})| \propto p({\bf x})^{2/3}.
\label{noisy_metric}
\end{equation}
The exponent accompanying the PDF in the Jacobian will be hereafter referred to as $\gamma$.

\subsection{Particular solutions for the response transform} \label{particular}

The above constraints on the Jacobian do not lead to a unique solution for the transform. It is well-known that independent responses from input signals following a certain PDF (the {\em infomax goal}) may be obtained in many different ways \cite {Hyvarinen99a}. A straightforward solution such as the equalization of the slices of the joint PDF at the input representation is not possible in practice from a finite set of samples due to the curse of dimensionality.

Iterative approaches related to Projection Pursuit~\cite{Huber85} may circumvent this problem. In fact, as pointed out in \cite {Laparra10rbig}, obtaining independent responses with deep neural networks is possible in many different ways: even using random rotations in the linear stages. However, in general, these iterative approaches lead to non-intuitive (or even meaningless) transform domains since $R$ is not constrained to preserve the local geometry of the input space. According to this non-uniqueness, the {\em infomax principle} and the nonlinear ICA goals are not enough to determine the sensors that reveal the intrinsic coordinates of data. Nevertheless, a wide range of unsupervised manifold learning techniques has been proposed to extract the latent coordinates from raw measurements, although not exactly in the context of nonlinear ICA.

Self-Organizing Maps~\cite{Kohonen82} and variants~\cite{Bishop98} are based on tuning a predefined topology in such a way that the nonlinearities of the sensors and the {\em complete} lattice of discrete responses are obtained simultaneously. These approaches are not feasible in highly dimensional situations since the number of nodes in the lattice explodes with dimensionality. Another group of techniques is based on the eigen-analysis of graphs and kernels related to the local structure of the data in the manifold~\cite{Tenenbaum2000,Scholkopf98,Weinberger04}, or on sparse matrices describing the local topology of the data~\cite{Roweis00,Belkin02}. Though efficient in many tasks, these {\em spectral} methods do not generally yield intuitive mappings between the original and the intrinsic curvilinear coordinates of the low dimensional manifold. In addition, even though a metric can be derived from particular kernel functions~\cite{Burges99}, the interpretation of the transformation is hidden behind implicit mappings, and out-of-sample extensions are typically difficult. An alternative family of manifold learning methods consider complicated manifolds as a mixture of local models~\cite{Kambhatla97} that are identified and conveniently merged into a single global representation~\cite{Roweis02,Verbeek02,Teh03,Brand03}. The explicit direct and inverse transforms to the intrinsic representation can be derived from the obtained mixture model.

Enforcing coordination between neighboring local models may be seen as reducing multi-information between variables in the coordination (or unfolding) operation. This relates NL-ICA with techniques based on coordination of local models. However, in~\cite{Roweis02,Verbeek02,Teh03,Brand03}, the effect of these local operations in the (eventually point-dependent) metric or line element was not explicitly analyzed. In the context of NL-ICA, an alternative way of merging locally disconnected representations was proposed in~\cite{Malo06b}. In that case, the global representation was based on the fact that the global NL-ICA at a certain point, $R({\bf x})$, may be differentially approximated by the local linear ICA separating matrix, $W({\bf x})$,~\cite{Lin99}. The issue was posed as an initial value problem and the global representation was obtained by integrating the local separating matrices in {\em arbitrary paths}. Note that, in the particular case of a mixture of Gaussians, the factorization of local models is consistent with (1) the Mahalanobis distance, and (2) the relation between the probability, the response and the metric under the noise-free infomax assumption\footnote{If the local models are assumed to be Gaussian, local factorization is achieved by local PCA and whitening. Specifically, if the local covariance can be decomposed as $\Sigma({\bf x})=B({\bf x}) \Lambda({\bf x}) B({\bf x})^\top$, the local separating matrix is just $W({\bf x})=\Lambda({\bf x})^{-1/2} B({\bf x})^\top$. In that case, the metric is $M({\bf x}) = W({\bf x})^\top \cdot W({\bf x}) = B({\bf x}) \Lambda({\bf x})^{-1} B({\bf x})^\top = \Sigma({\bf x})^{-1}$, i.e. the local Mahalanobis metric. Note also that $|\Sigma({\bf x})|^{-1/2}$ is inversely proportional to the volume of the local Gaussian support, thus, in this case, $|\nabla R({\bf x})|=|W({\bf x})| \propto p({\bf x})$ as in Eq.~\eqref{noise_free_jacobian}.}. However, the coordination by integrating the differential behavior in arbitrary paths as proposed in~\cite{Malo06b} only works for manifolds where the set of local basis functions fulfills the Stokes' theorem in the sense used in conservative vector fields. Moreover, the invertibility of the transform was not addressed therein \cite{Malo06b}.

In conclusion none of the above learning techniques is readily applicable to the simultaneous explanation of the non-linearities and adaptation of color vision mechanisms.

\subsection{Our proposal for the response transform}
\label{proposal}

The method proposed in Section \ref{generalSPCA} is based on the assumption of mixture of local models as classical methods based on vector quantization~\cite{Kambhatla97} and the variants that enforce model coordination~\cite{Roweis02,Verbeek02,Teh03,Brand03}. However, unlike in~\cite{Roweis02,Verbeek02,Teh03,Brand03}, no explicit mixture of models is computed in our approach. On the contrary, as in~\cite{Malo06b}, we propose to merge the local models by integrating some differential behavior, $\nabla R$. However, unlike~\cite{Malo06b}, the integration is done along a particular sequence of successive Principal Curves~\cite{Delicado01}, instead of using arbitrary paths. In this way, fulfillment of the Stokes' theorem in the manifold is no longer required, and a meaningful transformed domain is obtained since the differential behavior is integrated along meaningful trajectories in the manifold thus preserving the local topology of the input space.
Easy interpretation of the features defined by the Principal Curves solves the interpretability problem of nonlinear ICA techniques related to Projection Pursuit where the independent representation may be even random~\cite{Laparra10rbig}. Moreover, here we propose an explicitly tunable local metric according to the local PDF to achieve different goals such as \emph{infomax}, as in Eq.~\eqref{noise_free_metric}, or {\em error minimization}, as in Eq.~\eqref{noisy_metric}. Finally, the proposed transform is readily invertible which is a key issue to reproduce chromatic adaptation (see Section \ref{procedure}). Accordingly, the proposed response is suitable to reproduce the experimental facts reviewed in Section \ref{motivation} using the optimality criteria reviewed in Section \ref{sect_infomax}.


\section{Sequential Principal Curves Analysis (SPCA) with Local Metric}
\label{generalSPCA}

This section presents a manifold learning method, SPCA, that gives rise to an invertible transform, $R$. The technique can be seen as a method to design a set of eventually nonlinear sensors optimized according to the different goals reviewed in Section \ref{sect_infomax}. The method is first motivated by the particular characteristics of smooth manifolds. Then, we present the direct and inversion transforms, and finally study the impact of the metric on the solution.
\vspace{-0.3cm}
\subsection{Motivation}
\label{motivatingSPCA}
\begin{figure}[t!]
\begin{center}
\begin{tabular}{c}
\hspace{-0.25cm}
\includegraphics[width=14cm]{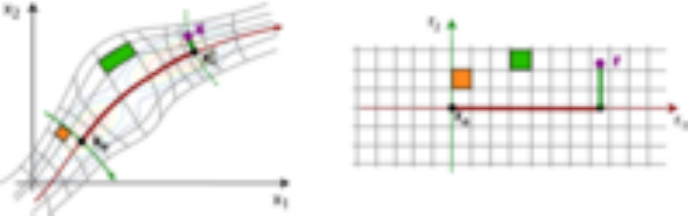}
\end{tabular}
\end{center}
\vspace{-0.45cm}\caption{\small Illustration of the SPCA with Local Metric. Left plot represents the input domain ${\bf x}$ and right plot represents the response domain ${\bf r}$. Colored contours represent the underlying PDF, $p({\bf x})$. The assumed curvilinear lattice (gray lines) is not explicitly computed, but just the path in bold curves. The proposed differential behavior (Eqs. \eqref{generalSPCA_jacobian} and \eqref{generalSPCA_metric}) implies that highly populated regions (such as the orange area) are expanded while lower density regions (such as the green area) are shrunk in the response domain (right figure). Given an origin of coordinates, ${\bf x}^o$, in the first PC (red line) and some point of interest, ${\bf x}$, the response for the point of interest is given by the lengths (the integrals in Eq.~\eqref{global_response2}) along the path consisting of successive PCs: the first (or standard) PC in red, and the second PC (in green) in the orthogonal subspace at ${{\bf x}_{\bot}^1}$, which is the (geodesic) orthogonal projection of ${\bf x}$ on the first PC.}
\label{digrama2}
\end{figure}

SPCA is based on the following characteristics of curved manifolds:
\vspace{-0.2cm}
\begin{enumerate}
\item Representing the data in curvilinear coordinates defined by Principal Curves (PCs) yields a representation where the data are unfolded. Intuitively, the dimensions become more meaningful in the sense that each one isolates a distinct feature of the signal (i.e. they are more independent). In~\cite{Malo11renips} it is shown that a set of local rotations and alignments along a PC reduces multi-information in curved manifolds. Specifically, unfolding along a PC is an adequate step towards independence since it makes equal the first moment of all the conditional PDFs along the curve.
\item Additional local processing after unfolding is required to achieve, either independence or minimum representation error, by using local expansions or compressions (i.e. locally changing the metric) of the unfolded domain.
\end{enumerate}
\vspace{-0.2cm}
The diagram in Fig.~\ref{digrama2} illustrates the intuitive ideas behind the proposed SPCA, that will be confirmed in the example of Fig.  \ref{gamba}. We assume the existence of a transformation defined by a curvilinear lattice made of recursively defined PCs along all dimensions. This lattice is similar to the topology assumed in Self-Organizing Maps (SOMs) and variants. In our context, each dimension of the lattice can be seen as the optimal feature for an eventually non-linear sensor in the original domain, defining a canonical direction in the transformed domain. However, unlike SOM, the whole lattice has not to be explicitly computed in order to find the transformation (response) of a particular point (stimulus). To do so, we propose to integrate a local differential behavior, $\nabla R(\mathbf{x})$, along a particular path.

The proposed integration path first follows the {\em standard PC of the set}~\cite{Hastie89} up to the geodesic projection of the point ${\bf x}$ on this {\em first} PC, ${{\bf x}_{\bot}^1}$.
The {\em first} PC provides a global summary of the whole dataset but the residual structure in the hyperplanes
orthogonal to this first PC may also be worth to be described.
In principle, any set of ($d-1$) linearly independent vectors living in the corresponding hyperplanes would be equally suited to form a linear basis to describe this structure. However, as noted by Delicado~\cite{Delicado01}, the structure
at those hyperplanes may also be nonlinear so it makes sense to draw secondary PCs to capture this residual structure.
These ideas imply that geodesic projections according to the local structure of the manifold can be obtained by secondary PCs~\cite{Malo11renips}.
After following the first PC, the path follows the {\em second PC}~\cite{Delicado01} at ${{\bf x}_{\bot}^1}$, i.e. the PC of the orthogonal subspace with regard to the first PC at the point ${{\bf x}_{\bot}^1}$. In this second segment, the path goes up to the geodesic projection of the point ${\bf x}$ on the second PC,  ${{\bf x}_{\bot}^2}$. This \emph{sequence} is continued until the last dimension.
The lengths of the curved segments represent the projections in each dimension of the new representation
and may be seen as the response of $d$ sensors tuned to curved features.

The proposed differential behavior, $\nabla R$, is based in the benefits of the joint consideration of unfolding and local equalization in smooth manifolds, and can be expressed as:
\begin{equation}
\nabla R({\bf x}) = D({\bf x}) \cdot \nabla U({\bf x}),
\label{generalSPCA_jacobian}
\end{equation}
where ${\bf u}=U({\bf x})$ is the unfolding transform that consists of concatenated local rotations along the proposed path made of a sequence of PCs, and the diagonal matrix $D({\bf x})$ represents the local length of the line element along this path (change of metric). Note that $\nabla U({\bf x})$ is orthonormal for all ${\bf x}$ since the unfolding $U$ can be formulated as a set of concatenated local rotations. In fact, in the selected method to draw one PC \cite{Malo11renips}, the curve consists of aligned rotations estimated by using local PCA. This is consistent with the fact that other PC algorithms use local PCA to estimate the tangent to the curves~\cite{Delicado01,Einbeck05}.

In order to adapt the metric to the density, we set the elements of $D$ using the marginal PDF on the unfolded coordinates and an appropriate exponent $\gamma\geq 0$:
\begin{equation}
     D({\bf u})_{ii} \propto p_{u_i}(u_i)^\gamma,
     \label{generalSPCA_scaling}
\end{equation}
where the marginal on each direction is estimated following $k$-neighborhood rule.
The metric induced in the input space is:
\begin{equation}
      M({\bf x}) =  \nabla U({\bf x})^\top \cdot D({\bf x})^2 \cdot \nabla U({\bf x}).
      \label{generalSPCA_metric}
\end{equation}
Assuming that local clusters can be factorized by the local rotations, and taking into account that $|\nabla U({\bf x})|=1$, we have
\begin{equation}
     |M({\bf x})| =  |D({\bf x})|^2 \propto \prod_{i=1}^{n} p_{u_i}(u_i)^{2\gamma} = p({\bf u})^{2\gamma} = p({\bf x})^{2\gamma} |\nabla U({\bf x})|^{-2\gamma} = p({\bf x})^{2\gamma},
     \label{determinant_SPCA_metric}
\end{equation}
which, with the appropriate choice of $\gamma$, is the behavior required in Eqs. \eqref{noise_free_metric} or \eqref{noisy_metric}.

\subsection{Direct transform}

Given an arbitrary origin of coordinates on the first PC, ${\bf x}^o$, assumed to give zero response, ${\bf r}^o = {\bf 0}$, and some point of interest, ${\bf x}$, the corresponding response is given by the following integration along the path on PCs described above (cf. Fig.~\ref{digrama2}):
\begin{equation}
{\bf r} = R({\bf x}) = C \cdot \int_{{\bf x}^o}^{{\bf x}} \nabla R({\bf x'}) \cdot d{\bf x'} = C \cdot \int_{{\bf x}^o}^{{\bf x}} D({\bf x'}) \cdot \nabla U({\bf x'}) \cdot d{\bf x'}, \\
\label{global_response1}
\end{equation}
where $C$ is just a constant diagonal matrix that independently scales each component of the response.
The selected path implies displacements in one PC at a time. According to this, in each segment of the path, the vectors $d{\bf u'} =\nabla U({\bf x'}) d{\bf x'}$ have only one non-zero component: the one corresponding to the considered PC at the considered segment. Therefore, the response of each sensor to the point ${\bf x}$ is just the length on each Principal Curve in the path from ${\bf x}^o$ to ${\bf x}$, measured according to the metric related to the local density with the selected exponent,
\begin{equation}
r_i = C_{ii} \cdot \int_{{\bf x}_{\bot}^{i-1}}^{{\bf x}_{\bot}^i} D({\bf x'}) \cdot \nabla U({\bf x'}) \cdot d{\bf x'} =  C_{ii} \int_{0}^{u_{i \bot}^i} p_{u_i}(u'_i)^\gamma \, du'_i, \label{global_response2}\\
\end{equation}

SPCA is initialized by setting (i) the origin of the coordinate system, and (ii) the scale of the different dimensions, $C_{ii}$, and
the order in which they will be visited by the sequential algorithm.
Sensible choices for the origin are those suggested in other bottom-up Principal Curve algorithms \cite{Delicado01,Einbeck05}:
the most dense point of the distribution (if known) or the mean of the data. Then a set
of $d$ locally orthogonal principal curves is drawn at the selected origin, which will be used to set the order and the relative scale of the dimensions.
In our case, we set the scaling constants $C_{ii}$ according to an information distribution criterion: we use the number of
quantization bins {\em per} dimension given by classical bit allocation results in transform coding~\cite{Gersho92}.
This is consistent with sorting the curvilinear dimensions according to the marginal entropy (higher entropy first).
Note that, other criteria could be used, as for instance the total standard deviation of the projected data (as in global PCA) or the total Euclidean length of the curvilinear axes.  

Once the dimensions have been sorted and the global scaling is set, SPCA obtains the transform of an arbitrary point ${\bf x}$ by \emph{sequentially} applying the next two steps. Step 1 traces a principal curve in the $i$-th direction from the previous starting point ${\bf x_{\bot}}^{i-1}$ (${\bf x_{\bot}}^{0}$ is the origin of the coordinate system).
Step 2 defines the line element in the drawn principal curve using the marginal PDF along the curve, $p_{u_i}(u_i)^\gamma$.
The response $r_i$ will be the integral of the line element from ${\bf x_{\bot}}^{i-1}$ to
the geodesic projection of ${\bf x}$ into the $i$-th principal curve, which is ${\bf x_{\bot}}^{i}$ (cf. Eq. \ref{global_response2}).
Details on the iterative refinement procedure to obtain the geodesic projections from orthogonal projections are given in \cite{Malo11renips}.
Since SPCA requires that the individual principal curves are drawn in particular directions from particular points,
appropriate algorithms to draw individual curves should operate in a bottom-up manner, as those in~\cite{Delicado01,Einbeck05}
or the particular one used here \cite{Malo11renips}.
The local-to-global behavior in the selected algorithm to draw each PC is necessary to identify the structure around
${\bf x_{\bot}}^{i-1}$ in the subspace locally orthogonal to the previous PC.

\begin{figure}[t!]
\small
\begin{center}
\hspace{-1.8cm}
\begin{tabular}{rccc}
& $\gamma = 0$ & $\gamma = 1$ & $\gamma =  1/3$\\
&\hspace{-0.2cm}\vspace{0.5cm}\includegraphics[width=5cm]{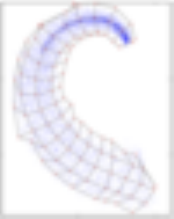} &
\hspace{-0.2cm}\includegraphics[width=5cm]{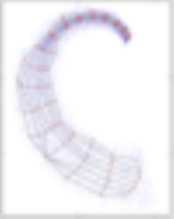} &
\hspace{-0.2cm}\includegraphics[width=5cm]{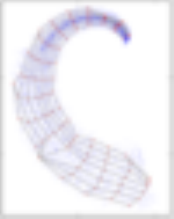} \\
& \hspace{-0.2cm} \vspace{-1.75cm}\includegraphics[width=5cm]{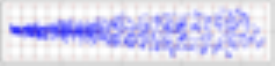} & & \\
& & \hspace{-0.2cm}\includegraphics[width=5cm,height=2.1cm]{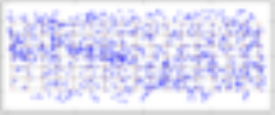} &
\hspace{-0.2cm}\includegraphics[width=5cm,height=2.1cm]{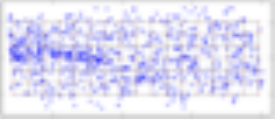} \\
\hline
RMSE &  0.55 & 0.56  &  0.53 \\
$MI$ (bits)  &  0.27  &  0.05  &  0.06 \\
\hline
\end{tabular}
\end{center}
\vspace{-0.25cm}
\caption{\small Infomax and error minimization through SPCA.
Samples of the sets in the first row were transformed using SPCA (second row) with different $\gamma$ value.
Additionally, Cartesian lattices in the response domain were inverted back into the input domain giving rise to the curved lattices in the top row.
Results are analyzed in terms of independence (Mutual Information), and reconstruction error (root-mean-square-error, RMSE). In each case, MI was computed in the corresponding transform domain, while RMSE values refer to the quantization error in the original domain using the corresponding lattices as codebook.
For the sake of reference, in the original domain results were $MI=0.75$ bits and RMSE=$0.63$ (using uniform scalar quantization).
Note how $\gamma = 1$ obtains better results in independence while $\gamma = \frac{1}{3}$ is better for RMSE minimization.}
\label{gamba}
\end{figure}

A Matlab implementation of SPCA with worked examples is available on-line\footnote{\texttt{http://isp.uv.es/spca.html}}.
Figure \ref{gamba} illustrates the performance of SPCA in a practical situation.

\subsection{Inverse transform}

A distinctive property of the method is the possibility of computing the inverse of the transform.
Given a set of samples from the same source, the origin in the input space, ${\bf x}^o$, and the scale and order of the dimensions, the computation of the inverse, ${\bf x}=R^{-1}({\bf r})$, is very simple. It involves drawing the first PC through the origin and taking the length $r_1$ on this curve, measured according to $p_{u_1}(u_1)^\gamma$. Displacement on the first curve by the length $r_1$ leads to the first projection ${{\bf x}_{\bot}^1}$. Then, the second locally orthogonal curve is drawn from ${{\bf x}_{\bot}^1}$, and one takes a second displacement $r_2$ on this second PC leading to the second projection, ${{\bf x}_{\bot}^2}$. This process is repeated sequentially in every dimension until the desired point ${\bf x}$ is found by taking the displacement $r_d$ from ${\bf x}^{d-1}_{\bot}$ on the {\em d}-th principal curve.

\subsection{Infomax and error minimization through SPCA}
\label{effect_of_parameters}

Here we present a synthetic experiment that stresses the usefulness of SPCA in sensor design, and study the effect of using different metrics ($\gamma = 0$, $1$ and $\frac{1}{3}$). We generated $10000$ samples from an illustrative curved manifold with changing PDF: half of the manifold has an increasing variance Laplacian distribution while the other half follows an increasing variance uniform distribution (Fig. \ref{gamba}).

The advantage of using Principal Curves to design a set of sensors is that their flexibility makes them suitable to describe curved manifolds, as pointed out in the example. No matter the metric used, an unwrapped representation of the data is obtained. When using $\gamma = 0$ the data is unfolded and the original local metric is preserved (e.g. the different distributions inside the manifold remain the same). When using $\gamma = 1$, we obtain a representation where the different distributions are almost uniformized leading to a representation where the different dimensions are almost independent.
Finally, when using  $\gamma = \frac{1}{3}$, the reconstruction error is minimized.
In this latter case, redundancy is certainly reduced with regard to the input domain, however, the kurtotic structure of the Laplacian is more visible in the transformed domain than in the second case.
Note also the differences in the distribution of the inverted lattices: while in the $\gamma = 0$ case, lattice cells are approximately uniform no matter the local
population (local metric independent of the PDF), in the other cases, the size is related to the population, e.g. the $\gamma = 1$ case results in tighter slices around the peak of the Laplacian. As anticipated in Section \ref{motivatingSPCA}, unfolding alone ($\gamma=0$) in general is not enough to remove redundancies, but additional processing, i.e. local changes in the metric related to the local PDF, are required to achieve independent components.


\section{Simulation of Color Psychophysics using SPCA}
\label{setup}

This section describes the procedure to simulate the experimental nonlinearities and the adaptation results described in Section \ref{motivation}
using SPCA on suitable ensembles of natural colors. Since the available experimental data involve adaptation under specific white and reddish illuminations, a new database was required.

\subsection{Database of calibrated natural color images}
\label{database1}

Calibrated measurements (tristimulus values instead of digital counts) and controlled white and reddish illumination on the same objects are needed to ensure the appropriate statistical adaptation conditions in the simulation of the psychophysics. Unfortunately, the current available color image databases do not fulfill such requirements because of different reasons:
\begin{enumerate}
\item Spectro-radiometric natural image databases, such as those used in~\cite{Brown94,Vrhel94,Webster97,Parraga98,Ruderman98,Nascimento02}, may be used to estimate the reflectance of natural surfaces under the flat Lambertian assumption. Then, these reflectance values can be used to obtain new tristimulus values under different illuminants. However, such procedure neglects the nonlinearities induced by geometric factors and mutual illumination, which are relevant factors to induce non-uniformities within the PDF support (as illustrated in Fig.~\ref{real_example}).
\item Databases where the illumination is carefully modified on the same objects, include spectro-radiometric examples~\cite{Brainard00} and uncalibrated examples~\cite{Barnard02,IJCV05}. The problem in these cases is that either the database consists of a very restricted set of artificial objects (unnatural clusters in the color space)~\cite{Brainard00,Barnard02}, or that the database is uncalibrated~\cite{Barnard02,IJCV05}.
\item Calibrated natural image color databases, such as~\cite{Doi03,Olmos04,Parraga09}, are not suitable for the simulation of color adaptation because either they do not include the same surfaces under the required controlled illuminants or they are not wide enough to find samples with the appropriate illumination.
\item A large database, such as that in~\cite{Funt03}, does include a wide range of scenes, a subset of which could match the desired white and reddish adaptation conditions but, unfortunately, it has been acquired with an uncalibrated video camera.
\end{enumerate}
These shortcomings led us to compile a new color image database of natural objects in controlled illumination conditions. We used a Macbeth light chamber equipped with standard CIE D65 and CIE A illuminants and we took the CIE XYZ pictures using a calibrated image colorimeter Lumicam1300. The accuracy of the illuminants and the measurements was checked by taking pictures of 10 hue pages of the Munsell's Book of Color and comparing the results with theoretical tristimulus values computed from the reflectance of the samples and the radiances of the illuminants. The database consists of 75 scenes of size $1000\times1280$. For each scene, two pictures were taken under CIE A and CIE D65 illuminants. The scenes include plants and flowers, natural terrain and materials, samples of colored fabric, office material, and Munsell chips.
The database is publicly available on-line\footnote{\texttt{http://isp.uv.es/data\_color.htm}} and it is suitable for other accurate experiments on color constancy and chromatic adaptation. Details on the experimental procedure to gather the database are given in the dedicated web site. 

\begin{figure}[t!]
\begin{center}
\setlength{\tabcolsep}{2pt}
\begin{tabular}{ccccc|ccccc}
\multicolumn{5}{c}{CIE D65 illuminant} & \multicolumn{5}{c}{CIE A illuminant} \\
\includegraphics[height=1.4cm,width=1.4cm]{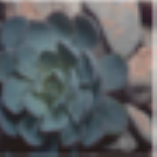} &
\includegraphics[height=1.4cm,width=1.4cm]{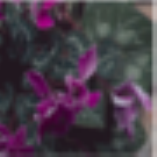} &
\includegraphics[height=1.4cm,width=1.4cm]{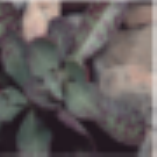} &
\includegraphics[height=1.4cm,width=1.4cm]{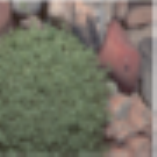} &
\includegraphics[height=1.4cm,width=1.4cm]{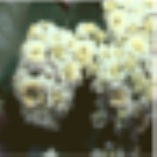} &
\includegraphics[height=1.4cm,width=1.4cm]{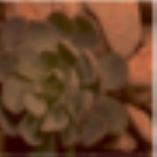} &
\includegraphics[height=1.4cm,width=1.4cm]{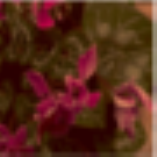} &
\includegraphics[height=1.4cm,width=1.4cm]{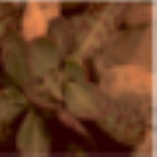} &
\includegraphics[height=1.4cm,width=1.4cm]{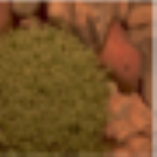} &
\includegraphics[height=1.4cm,width=1.4cm]{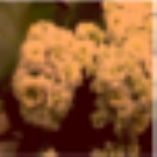} \\
\includegraphics[height=1.4cm,width=1.4cm]{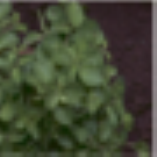} &
\includegraphics[height=1.4cm,width=1.4cm]{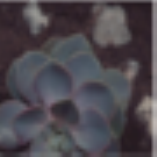} &
\includegraphics[height=1.4cm,width=1.4cm]{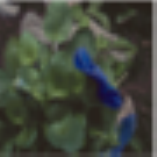} &
\includegraphics[height=1.4cm,width=1.4cm]{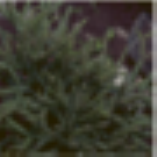} &
\includegraphics[height=1.4cm,width=1.4cm]{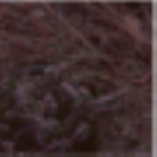} &
\includegraphics[height=1.4cm,width=1.4cm]{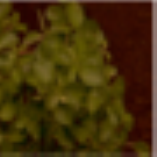} &
\includegraphics[height=1.4cm,width=1.4cm]{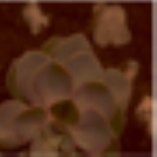} &
\includegraphics[height=1.4cm,width=1.4cm]{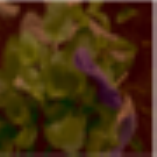} &
\includegraphics[height=1.4cm,width=1.4cm]{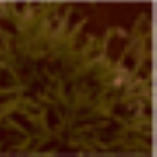} &
\includegraphics[height=1.4cm,width=1.4cm]{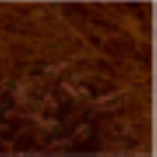} \\
\includegraphics[height=1.4cm,width=1.4cm]{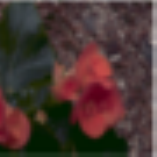} &
\includegraphics[height=1.4cm,width=1.4cm]{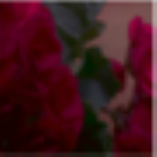} &
\includegraphics[height=1.4cm,width=1.4cm]{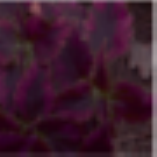} &
\includegraphics[height=1.4cm,width=1.4cm]{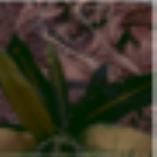} &
\includegraphics[height=1.4cm,width=1.4cm]{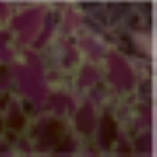} &
\includegraphics[height=1.4cm,width=1.4cm]{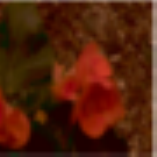} &
\includegraphics[height=1.4cm,width=1.4cm]{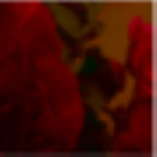} &
\includegraphics[height=1.4cm,width=1.4cm]{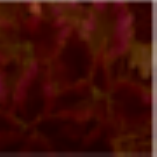} &
\includegraphics[height=1.4cm,width=1.4cm]{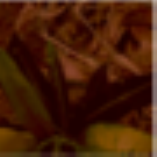} &
\includegraphics[height=1.4cm,width=1.4cm]{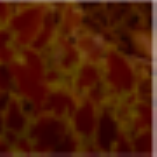} \\
\includegraphics[height=1.4cm,width=1.4cm]{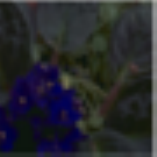} &
\includegraphics[height=1.4cm,width=1.4cm]{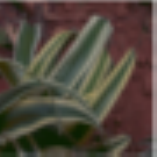} &
\includegraphics[height=1.4cm,width=1.4cm]{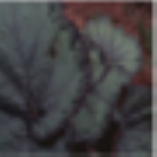} &
\includegraphics[height=1.4cm,width=1.4cm]{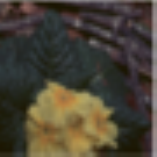} &
\includegraphics[height=1.4cm,width=1.4cm]{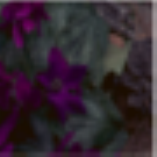} &
\includegraphics[height=1.4cm,width=1.4cm]{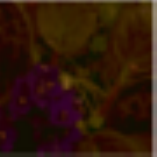} &
\includegraphics[height=1.4cm,width=1.4cm]{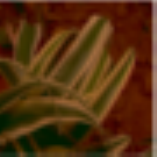} &
\includegraphics[height=1.4cm,width=1.4cm]{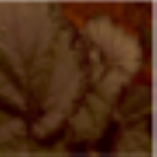} &
\includegraphics[height=1.4cm,width=1.4cm]{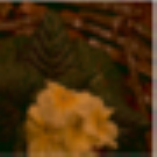} &
\includegraphics[height=1.4cm,width=1.4cm]{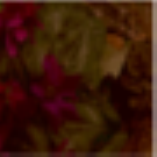} \\
\includegraphics[height=1.4cm,width=1.4cm]{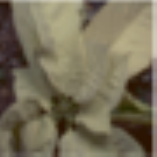} &
\includegraphics[height=1.4cm,width=1.4cm]{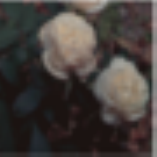} &
\includegraphics[height=1.4cm,width=1.4cm]{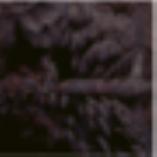} &
\includegraphics[height=1.4cm,width=1.4cm]{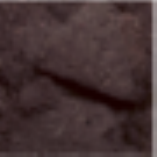} &
\includegraphics[height=1.4cm,width=1.4cm]{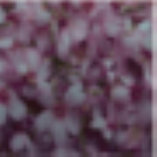} &
\includegraphics[height=1.4cm,width=1.4cm]{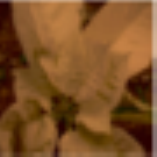} &
\includegraphics[height=1.4cm,width=1.4cm]{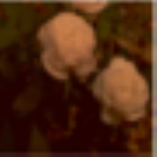} &
\includegraphics[height=1.4cm,width=1.4cm]{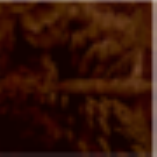} &
\includegraphics[height=1.4cm,width=1.4cm]{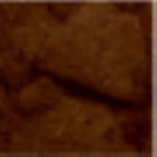} &
\includegraphics[height=1.4cm,width=1.4cm]{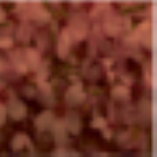} \\
\includegraphics[height=1.4cm,width=1.4cm]{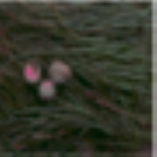} &
\includegraphics[height=1.4cm,width=1.4cm]{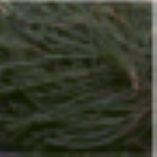} &
\includegraphics[height=1.4cm,width=1.4cm]{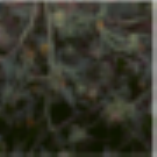} &
\includegraphics[height=1.4cm,width=1.4cm]{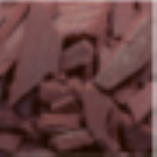} &
\includegraphics[height=1.4cm,width=1.4cm]{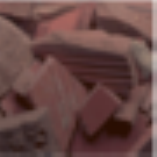} &
\includegraphics[height=1.4cm,width=1.4cm]{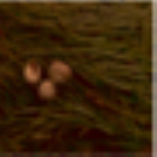} &
\includegraphics[height=1.4cm,width=1.4cm]{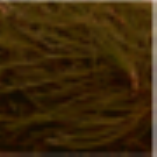} &
\includegraphics[height=1.4cm,width=1.4cm]{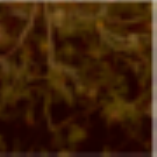} &
\includegraphics[height=1.4cm,width=1.4cm]{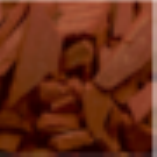} &
\includegraphics[height=1.4cm,width=1.4cm]{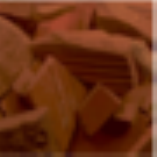} \\
\includegraphics[height=1.4cm,width=1.4cm]{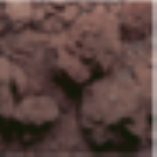} &
\includegraphics[height=1.4cm,width=1.4cm]{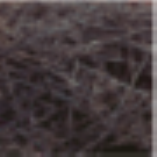} &
\includegraphics[height=1.4cm,width=1.4cm]{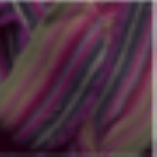} &
\includegraphics[height=1.4cm,width=1.4cm]{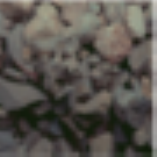} &
\includegraphics[height=1.4cm,width=1.4cm]{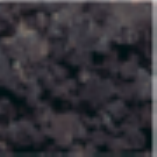} &
\includegraphics[height=1.4cm,width=1.4cm]{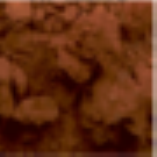} &
\includegraphics[height=1.4cm,width=1.4cm]{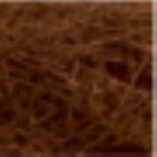} &
\includegraphics[height=1.4cm,width=1.4cm]{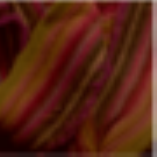} &
\includegraphics[height=1.4cm,width=1.4cm]{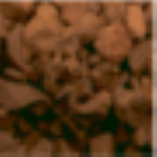} &
\includegraphics[height=1.4cm,width=1.4cm]{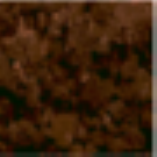} \\
\includegraphics[height=1.4cm,width=1.4cm]{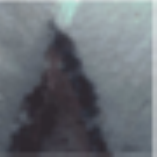} &
\includegraphics[height=1.4cm,width=1.4cm]{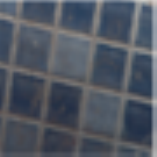} &
\includegraphics[height=1.4cm,width=1.4cm]{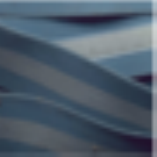} &
\includegraphics[height=1.4cm,width=1.4cm]{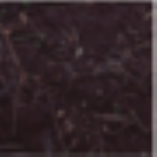} &
\includegraphics[height=1.4cm,width=1.4cm]{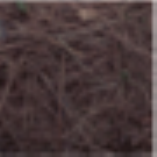} &
\includegraphics[height=1.4cm,width=1.4cm]{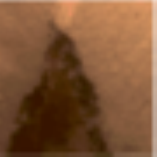} &
\includegraphics[height=1.4cm,width=1.4cm]{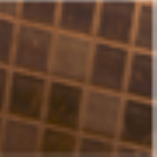} &
\includegraphics[height=1.4cm,width=1.4cm]{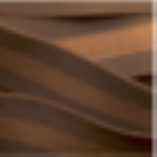} &
\includegraphics[height=1.4cm,width=1.4cm]{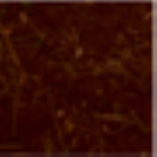} &
\includegraphics[height=1.4cm,width=1.4cm]{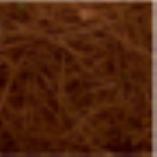} \\
\includegraphics[height=1.4cm,width=1.4cm]{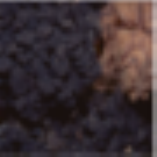} &
\includegraphics[height=1.4cm,width=1.4cm]{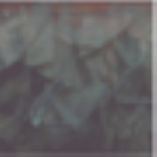} &
\includegraphics[height=1.4cm,width=1.4cm]{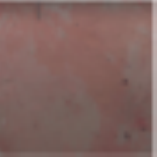} &
\includegraphics[height=1.4cm,width=1.4cm]{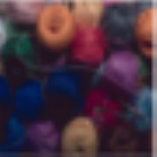} &
\includegraphics[height=1.4cm,width=1.4cm]{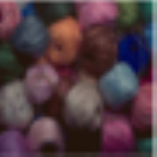} &
\includegraphics[height=1.4cm,width=1.4cm]{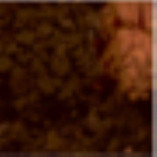} &
\includegraphics[height=1.4cm,width=1.4cm]{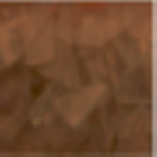} &
\includegraphics[height=1.4cm,width=1.4cm]{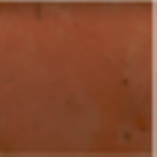} &
\includegraphics[height=1.4cm,width=1.4cm]{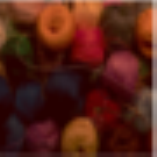} &
\includegraphics[height=1.4cm,width=1.4cm]{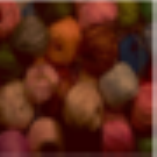} \\
\includegraphics[height=1.4cm,width=1.4cm]{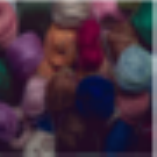} &
\includegraphics[height=1.4cm,width=1.4cm]{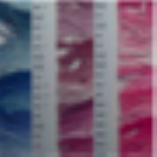} &
\includegraphics[height=1.4cm,width=1.4cm]{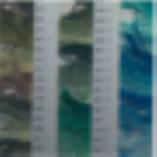} &
\includegraphics[height=1.4cm,width=1.4cm]{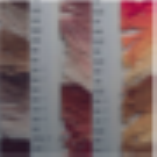} &
\includegraphics[height=1.4cm,width=1.4cm]{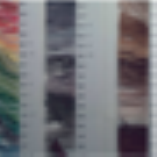} &
\includegraphics[height=1.4cm,width=1.4cm]{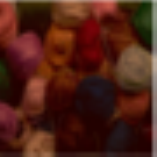} &
\includegraphics[height=1.4cm,width=1.4cm]{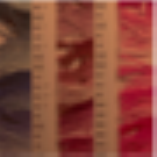} &
\includegraphics[height=1.4cm,width=1.4cm]{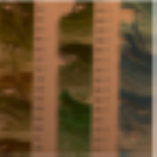} &
\includegraphics[height=1.4cm,width=1.4cm]{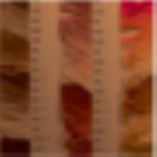} &
\includegraphics[height=1.4cm,width=1.4cm]{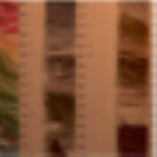} \\
\end{tabular}
\end{center}
\vspace{-0.45cm}\caption{\small Scenes used in the statistical simulations under different illumination.}
\label{database}
\end{figure}

In our specific experiments, we used 50 images excluding the Munsell chips and the pictures of the (too flat) artificial objects. This amounts to $64\cdot10^6$ color samples for each illumination. Figure \ref{database} shows the pictures considered in our experiments. Transformation from CIE XYZ values to VGA digital counts for visualization purposes in Fig.~\ref{database} was done using standard display calibration data~\cite{Colorlab00}. This may introduce some color reproduction errors in Fig.~\ref{database}. However, note that these eventual errors do not affect the simulations, which were done from the raw CIE XYZ measurements.

\subsection{Procedure for the simulation of color mechanisms behavior using SPCA}
\label{procedure}

\begin{figure}[t!]
\begin{center}
\begin{tabular}{ccc}
\includegraphics[width=4cm]{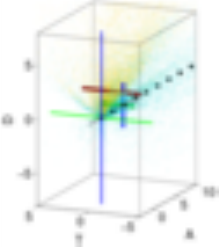} &
\includegraphics[width=3cm]{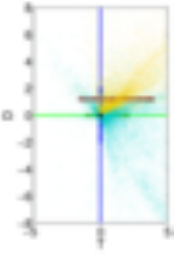} &
\includegraphics[width=4.3cm]{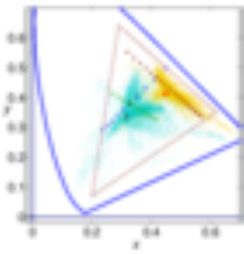} \\
\includegraphics[width=4cm]{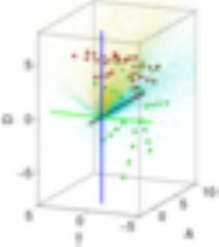} &
\includegraphics[width=3cm]{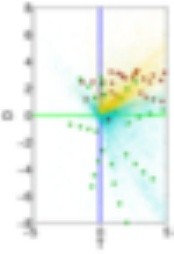} &
\includegraphics[width=4.3cm]{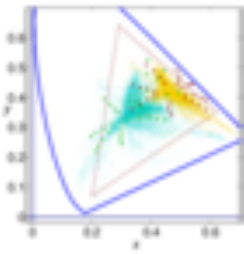} \\
\end{tabular}
\end{center}
\vspace{-0.45cm}\caption{\small Training and test points for the simulation of the psychophysics.
Small points represent the training samples of the database (cyan for CIE D65 and yellow for CIE A) and large points
represent the different sets of test points. In these plots only a subset of randomly selected training points is shown
for better visualization.
\textbf{Top:} Black, green and blue dots in the top row are the considered points to simulate the nonlinearity of the A, T, and D mechanisms respectively. In these cases the response is computed using the CIE D65 training set (cyan dots). Dark red dots are the considered points to simulate the nonlinearity in a reddish environment using the CIE A training set (yellow dots). Top left plot shows the data in the tristimulus ATD space. A zoom around the origin is shown here for better visualization. However note that test points along the achromatic axis spread up to A = 80. Top center plot show the data in the (T, D) plane. Top right plot shows the data in the CIE xy chromaticity diagram. In this case, the chromatic coordinates of the CIE D65 and CIE A illuminants are also shown for reference (larger gray and red dots respectively). \textbf{Bottom:} Green and red points in the bottom row are the considered points to compute the corresponding pairs using the CIE D65 training set (cyan dots) and the CIE A training set (yellow dots).
}
\label{data_exp1}
\end{figure}

\paragraph{Simulation of nonlinearities.} Nonlinearities along the A, T and D dimensions of the color space and their variations under adaptation changes can be reproduced by computing the response of the SPCA mechanisms on the corresponding axes and the appropriate adaptation environment (CIE D65 set or CIE A set). Figure \ref{data_exp1} (top row) shows the points considered in the simulation in the Ingling and Tsou ATD space. This space is selected as the input linear representation instead of the MacLeod and Boynton ATD space~\cite{MacLeod79} used in~\cite{Gegenfurtner92} because it better reproduces basic psychophysical data such as color matching functions similar to Jameson and Hurvich hue cancellation curves, and appropriate orientation of the McAdam's ellipse at the white point~\cite{Capilla98}.

Experimental results on nonlinearities can be simulated in two different ways that we will refer to as the {\em psychophysical paradigm} and the {\em physiological paradigm}. In the {\em physiological paradigm}, we assume we have access to the response of each mechanism as in an ideal neuron recording. In this case, we can register the response of the corresponding sensor (the first sensor in the A case, the second in the T case, and the third in the D case), and we can simulate the incremental thresholds of these mechanisms from the derivative (slope) of the responses at the considered points. In the {\em psychophysical paradigm}, the isolated responses are assumed to be inaccessible. On the contrary, we assume a certain summation of the variations in the responses of the sensors (e.g. the Euclidean norm for simplicity). The incremental threshold is reached when this norm achieves some prefixed value. In this way, we can simulate the thresholds and, by integrating their inverse, the underlying response can be derived as in psychophysics, cf. Eq.~\eqref{Fechner}.

As in any finite color database, a certain bias is expected~\cite{Koenderink10}. Figure \ref{data_exp1} displays the existing bias in the collected database. Note that the maximum of the PDFs (the statistical adaptation points) in each case do not match the CIE D65 and the CIE A chromaticities due to the particular objects in the database. Moreover, the most dense points are also shifted from the origin in the considered linear representation (intersection point between the T and D axes in the top right plot in Fig.~\ref{data_exp1}). This will introduce the corresponding bias in the results but it does not reduce the generality of the results, as recognized in previous statistical studies also dealing with biased databases~\cite{MacLeod03,MacLeod03b}.

\paragraph{Simulation of adaptation.}

\begin{figure}[t!]
\vspace{2.5cm}
\begin{center}
\begin{tabular}{cc}
\hspace{-0.25cm}
& \vspace{0cm} SPCA solution\\
& \vspace{-6cm} \includegraphics[height=3cm,width=3cm]{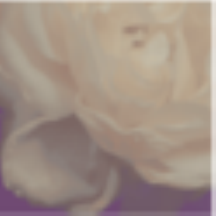}\\
\includegraphics[width=12cm]{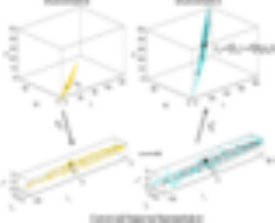} & \\
\end{tabular}
\end{center}
\vspace{-0.5cm}
\caption{\small Dataset shift compensation by using the corresponding pair concept and SPCA. Transforms leading to the corresponding latent coordinates of the manifold in the environments $A$ and $B$ may be used to estimate the position in environment $A$ of new points measured in environment $B$. Unlike in the linear adaptation cases in Fig. \ref{real_example}, the proposed nonlinear transform not only removes the yellowish appearance, but additionally the shadows are reduced as expected in a better PDF matching. In particular, note how the highlighted point $\mathbf{x}_B$ (the same one as in Fig. \ref{real_example}) results in a white, \emph{higher luminance} corresponding point $\mathbf{\hat{x}}_A$.}
\label{adaptation}
\end{figure}

Our proposal for domain adaptation using SPCA as response transform, $R$, is inspired by the {\em corresponding pair procedure} framework used in chromatic adaptation models to predict corresponding stimuli~\cite{Capilla04}. In this framework, linear measurements (e.g. CIE XYZ, LMS or ATD tristimulus values, ${\bf x}$), obtained in different conditions, $C$, are transformed according to an invertible color appearance model described by a transform, $R$, to a canonical space, e.g. the space of perceptual descriptors, ${\bf r}$, related to brightness, hue and colorfulness. The direct and inverse transforms depend on the measurement conditions, $C$. Measurement conditions may include information about the environment (e.g. spectral illumination, geometry), or information about the properties of the measurement system (e.g. normal or defective observers):
\begin{equation}
{\bf r} = R_C({\bf x}).
\end{equation}
Once a given point acquired in situation $B$ is transformed to the canonical representation of perceptual descriptors, it can be transformed back into the input domain of situation $A$ by using the inverse of the transform for situation $A$ (see Eq. (1) in~\cite{Capilla04}):
\begin{equation}
{\bf \widehat{x}_A} = R_A^{-1}({\bf r_B}) = R_A^{-1}(R_B({\bf x_B})).
\label{corresponding_eq}
\end{equation}
In problems where changes in the PDF due to non-interesting sources are smooth such as the ones found in color vision, we conjecture that transforms to canonical domains defined by the meaningful latent variables of the manifold can be used to solve the dataset shift problem. Changes in the PDF may give rise to nonlinear deformations of the curvilinear coordinates of the manifold and to changes on the length measures on them.
By using the technique proposed in Section \ref{generalSPCA}, one should be able to arrive to the same response thus achieving a canonical invariant representation. The results in Fig.~\ref{adaptation} illustrate this concept.

In the simulations of the empirical chromatic adaptation data, we transform one of the sets of the corresponding color data (e.g. colors at the bottom row in Fig.~\ref{data_exp1}) using the learned transform with SPCA with the appropriate adaptation environment (e.g. the CIE D65 training set). Then, the obtained responses are inverted back into the ATD space by using the inverse SPCA with the other environment set (e.g. the CIE A training set). In our simulations, the procedure was applied in both directions: from CIE D65 to CIE A, and viceversa. In each case, the computed colors have to be compared with the experimental data in Fig.~\ref{non_linearities} (bottom row). Figure \ref{data_exp1} (bottom row) shows the training and test data for the corresponding pairs experiment.


\section{Numerical results and Discussion}
\label{nonlinear_results}

This section shows how both nonlinearity and adaptation phenomena emerge from tristimulus samples using the proposed SPCA.
In particular, we show the results for the nonlinear behavior along the ATD directions and the corresponding data reproduction
using SPCA with the {\em error minimization} and the {\em infomax} strategies (exponents $\gamma=1/3$ and $\gamma=1$ respectively).

\paragraph{Parameters for drawing a principal curve.}
The parameters associated to the particular algorithm used to draw individual Principal Curves refer to the rigidity of the assumed underlying grid, or equivalently, to the freedom to find curved axes far from the global linear PCA solution. In our implementation, rigidity is controlled with the locality $k$, the step size $\tau$, and the stiffness $q$ (details in \cite{Malo11renips}). In the color statistics problem, the manifold is not globally curved and changes in spectral illumination induce almost linear rotations. In this situation, the relevant nonlinearities come from the non-uniform data distribution inside the PDF support basically due to the statistics of reflectance and geometric issues such as oblique illumination (cf. Fig.~\ref{real_example}). These non uniformities are not taken into account by the rigidity parameters of the particular PC algorithm ($k$, $\tau$ and $q$) but by the non-Euclidean metric used in the SPCA framework (i.e. by the {\em infomax} or the {\em error minimization} strategies). According to this, in the problem at hand, the relevant comparison is between these strategies, which incidentally is the biologically interesting issue.

In our case, the rigidity constraints of the principal curves algorithm have been optimized for best performance and applied in the same way in
both {\em infomax} and {\em error minimization} cases. Optimization of rigidity parameters has been done by exhaustive search in a discrete grid in the parameter space. The best values found were: $k=0.2$ (20 \% of the samples in the neighborhoods), $\tau=15$ in Euclidean units in the considered ATD space, and $q=16$ for the stiffness parameter. These parameters imply assuming a relatively rigid underlying grid, which makes sense in the color statistics problem.

\paragraph{Results.}

\begin{figure}[t!]
\begin{center}
\begin{tabular}{ccc}
\includegraphics[height=3.3cm,width=4cm]{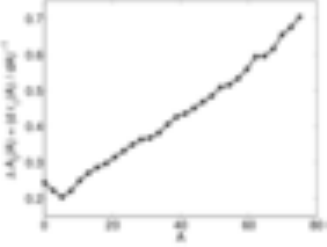} &
\includegraphics[height=3.3cm,width=4cm]{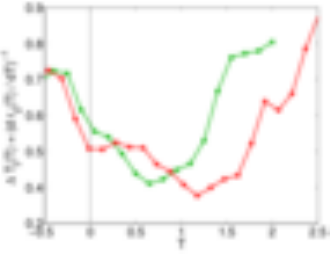} &
\includegraphics[height=3.3cm,width=4cm]{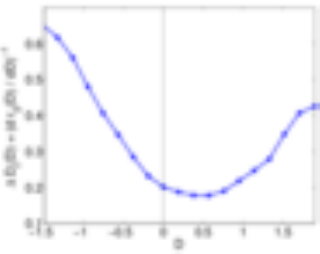} \\
\includegraphics[height=3.3cm,width=4cm]{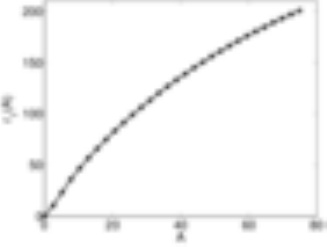} &
\includegraphics[height=3.3cm,width=4cm]{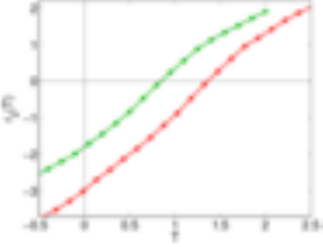} &
\includegraphics[height=3.3cm,width=4cm]{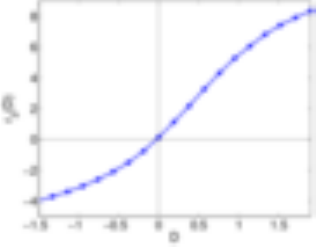} \\
\end{tabular}
\begin{tabular}{cc}
\includegraphics[height=4cm,width=4cm]{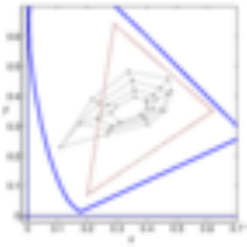} &
\includegraphics[height=4cm,width=4cm]{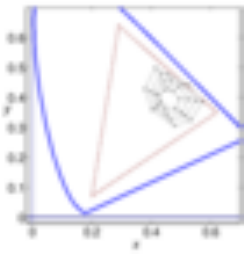}\\
\end{tabular}
\end{center}
\vspace{-0.45cm}\caption{\small Simulation of psychophysics with SPCA using the {\em error minimization} strategy.}
\label{SPCA_results_tercio}
\end{figure}

\begin{figure}[t!]
\begin{center}
\begin{tabular}{ccc}
\includegraphics[height=3.3cm,width=4cm]{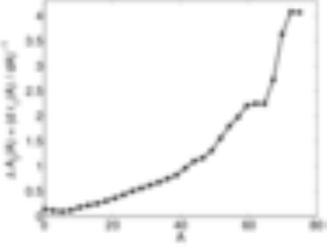} &
\includegraphics[height=3.3cm,width=4cm]{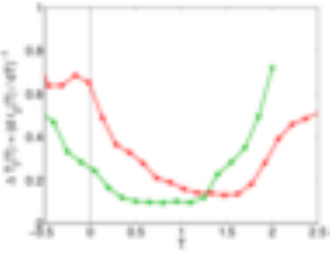} &
\includegraphics[height=3.3cm,width=4cm]{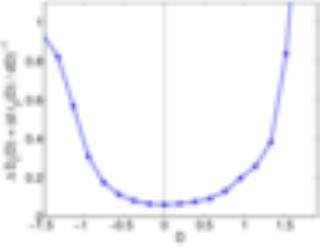} \\
\includegraphics[height=3.3cm,width=4cm]{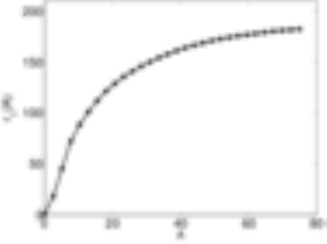} &
\includegraphics[height=3.3cm,width=4cm]{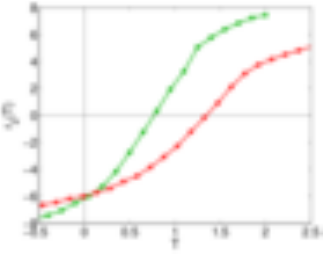} &
\includegraphics[height=3.3cm,width=4cm]{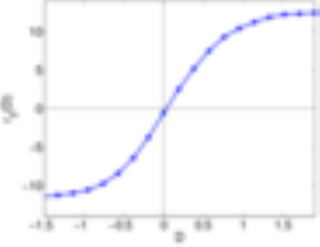} \\
\end{tabular}
\begin{tabular}{cc}
\includegraphics[height=4cm,width=4cm]{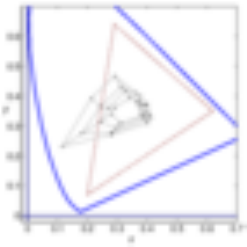} &
\includegraphics[height=4cm,width=4cm]{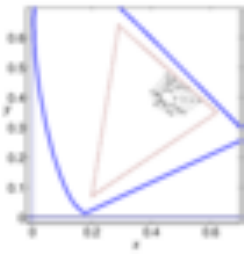}\\
\end{tabular}
\end{center}
\vspace{-0.45cm}\caption{\small Simulation of psychophysics with SPCA using the {\em infomax} strategy.}
\label{SPCA_results}
\end{figure}

Figures \ref{SPCA_results_tercio} and \ref{SPCA_results} show the SPCA results for the {\em error minimization} and the {\em infomax} strategies respectively. In the reproduction of the thresholds and nonlinearities, we used both the {\em physiological} and the {\em psychophysical} paradigms. Results are very similar for both paradigms. We just show the {\em physiological-like} result in each case. Black lines in the plots indicate the axes in the input ATD space. The deviation of the responses from the origin comes from the bias of the database. This just stresses the fact that the algorithm is adapted to the environment represented by the PDFs, which are biased with regard to the particular adaptation conditions used in the experiments. As stated above, this kind of bias does not represent a failure of the model, but the (necessarily) restricted nature of the database~\cite{MacLeod03,MacLeod03b}.

\begin{figure}[t!]
\begin{center}
\begin{tabular}{ccc}
\includegraphics[height=3.3cm,width=4cm]{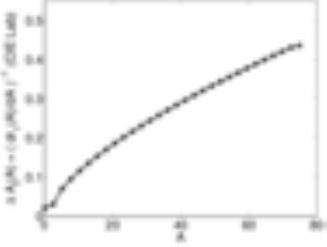} &
\includegraphics[height=3.3cm,width=4cm]{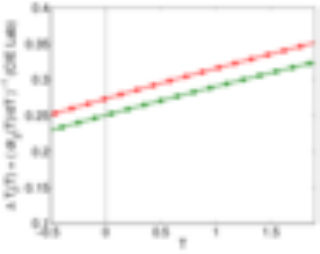} &
\includegraphics[height=3.3cm,width=4cm]{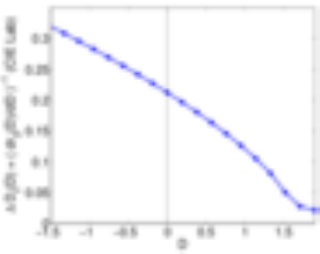} \\
\includegraphics[height=3.3cm,width=4cm]{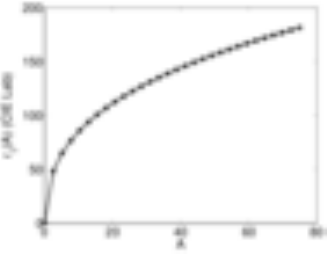} &
\includegraphics[height=3.3cm,width=4cm]{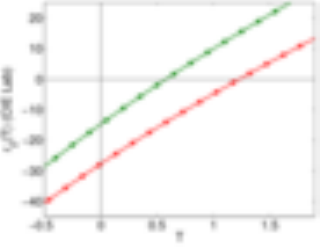} &
\includegraphics[height=3.3cm,width=4cm]{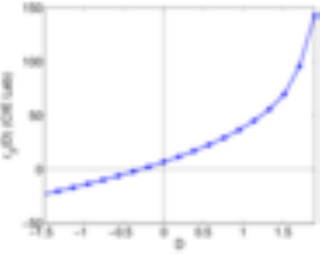} \\
\end{tabular}
\begin{tabular}{cc}
\includegraphics[height=4cm,width=4cm]{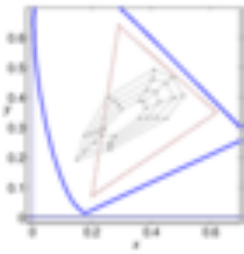} &
\includegraphics[height=4cm,width=4cm]{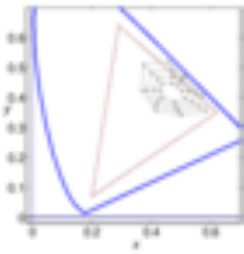}\\
\end{tabular}
\end{center}
\vspace{-0.45cm}\caption{\small Simulation of psychophysics with CIELab color appearance model.}
\label{CIElab_results}
\end{figure}

\begin{figure}[t!]
\begin{center}
\begin{tabular}{ccc}
\includegraphics[height=3.3cm,width=4cm]{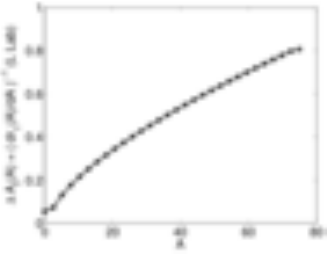} &
\includegraphics[height=3.3cm,width=4cm]{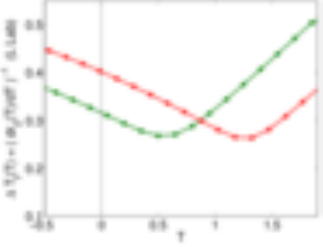} &
\includegraphics[height=3.3cm,width=4cm]{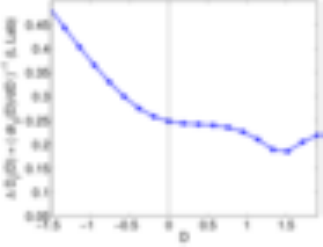} \\
\includegraphics[height=3.3cm,width=4cm]{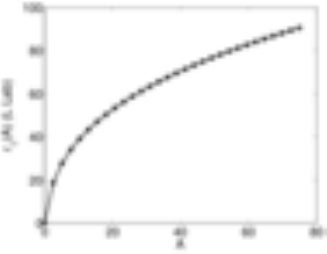} &
\includegraphics[height=3.3cm,width=4cm]{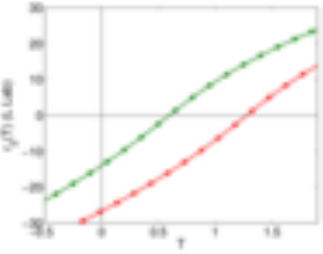} &
\includegraphics[height=3.3cm,width=4cm]{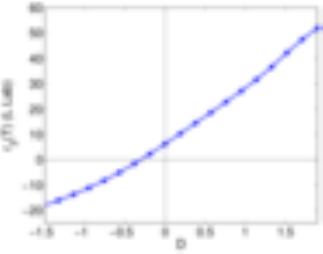} \\
\end{tabular}
\begin{tabular}{cc}
\includegraphics[height=4cm,width=4cm]{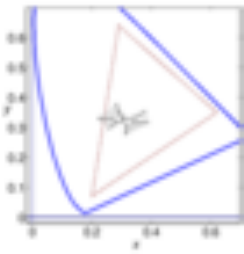} &
\includegraphics[height=4cm,width=4cm]{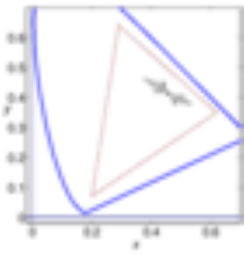}\\
\end{tabular}
\end{center}
\vspace{-0.45cm}\caption{\small Simulation of psychophysics with LLAb  color appearance model.}
\label{LLab_results}
\end{figure}

\begin{figure}[t!]
\begin{center}
\begin{tabular}{ccc}
\includegraphics[height=3.3cm,width=4cm]{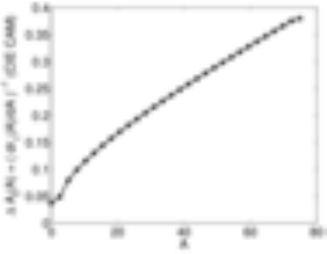} &
\includegraphics[height=3.3cm,width=4cm]{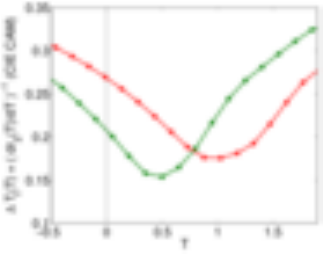} &
\includegraphics[height=3.3cm,width=4cm]{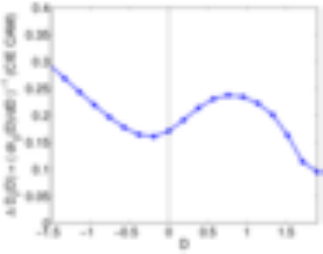} \\
\includegraphics[height=3.3cm,width=4cm]{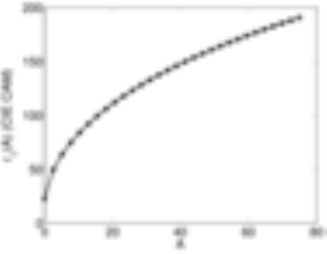} &
\includegraphics[height=3.3cm,width=4cm]{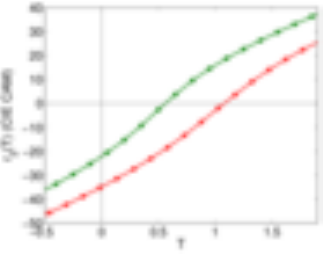} &
\includegraphics[height=3.3cm,width=4cm]{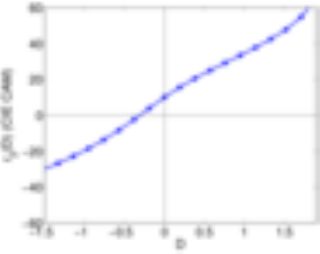} \\
\end{tabular}
\begin{tabular}{cc}
\includegraphics[height=4cm,width=4cm]{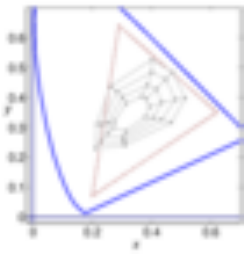} &
\includegraphics[height=4cm,width=4cm]{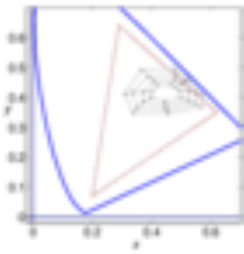}\\
\end{tabular}
\end{center}
\vspace{-0.45cm}\caption{\small Simulation of psychophysics with CIECAM color appearance model.}
\label{CIEcam_results}
\end{figure}

As convenient reference to assess the quality of the statistical results, which use no perceptual information, we also explored the performance of several psychophysically-based Color Appearance Models with the required elements (chromatic adaptation transform and nonlinearities in opponent channels): CIELab~\cite{Robertson77}, SVF~\cite{Seim86}, RLAB~\cite{Fairchild96}, LLab~\cite{Luo96}. See~\cite{Fairchild05} for a recent collective comparison of these models. A Matlab implementation of the considered color appearance models is available on-line \cite{Colorlab00}. Figures \ref{CIElab_results}, \ref{LLab_results} and \ref{CIEcam_results} show the results for CIELab, LLab and CIECAM, respectively. These results illustrate the general trend when using empirical models and stress the challenge represented by the simultaneous reproduction of nonlinearities and color adaptation data: widely used traditional models such as CIELab, LLab, RLab and SVF fail to simultaneously reproduce both aspects of the phenomenology. They reproduce either the color adaptation (as in the CIELab case) or the nonlinear behavior (as in the LLab case). Only the more recent CIECAM model is able to approximately account for both psychophysical aspects.

Interestingly, the results show that both SPCA strategies ({\em error minimization} and {\em infomax}) qualitatively reproduce the trends in both aspects of the phenomenology. SPCA gives rise to nonlinear responses in the ATD directions that shift in the appropriate way when changing the adaptation environment from white to reddish illumination. Moreover, SPCA qualitatively reproduces the shift in the corresponding colors and the orientation of chroma circles, both in the CIE D65 from CIE A data and viceversa. This general behavior comes from the fact that the proposed algorithm follows the changes in the PDFs, and has increased resolution, higher sensitivity, in the more populated regions (as illustrated in the example of Fig. \ref{gamba}). Note that the minima in the thresholds in the T and D axes (Figs. \ref{SPCA_results_tercio} and \ref{SPCA_results}, top row) coincide with the corresponding maximum in each PDF (Fig.~\ref{data_exp1}).

\paragraph{Discussion.}
The proposed statistical explanation is more general than previous statistical approaches only focused on one of the two phenomena. On the one hand, PCA-based approaches such as~\cite{Atick93,Webster97} do reproduce the shift and scaling of Luo et al.
corresponding colors (results not shown), but their linear nature implies that they cannot reproduce the nonlinearities in ATD. We did not check the performance of more recent linear-ICA-based approaches such as~\cite{Sejnowski01,Doi03} in reproducing corresponding colors, but in any case, they inherently suffer from the same limitation with regard to the Weber's Law and the chromatic nonlinearities. On the other hand, Laughlin and MacLeod et al. certainly introduced strategies to account for the nonlinearities~\cite{Laughlin83,Twer01,MacLeod03,MacLeod03b} but they did not explicitly propose a multidimensional transform to perform the analysis, so their ideas cannot be used to straightforwardly derive the corresponding colors dataset.

The performance of our non-analytic technique is consistent with the general conclusions found by Abrams et al.~\cite{Abrams07}
where discrimination and color constancy are simultaneously considered. They found that analytic models based on Von-Kries adaptation, color opponent transforms and dimension-wise nonlinearities can be simultaneously optimal in discrimination and adaptation under spectral illumination changes, but not when the reflectance ensemble is substantially changed. Here we did not try to address the optimality of the proposed technique in terms of ROC as in~\cite{Abrams07}, but it is obvious by the construction of SPCA that compensation of observation conditions is not going to be possible for our technique if the objects giving rise to the different adaptation ensembles are very different from each other. Our technique needs wide enough reflectance ensembles for a proper adaptation: if one tristimulus manifold comes from a wide set of objects while the other comes from a (restricted) set of, say, mainly reddish objects, the manifolds do not qualitatively match, so color compensation results are not going to be accurate. In Abrams et al. terms, a different set of parameters (a different mechanism) is needed in this situation. Our results represent a data-driven alternative to Abrams et al. approach since, in our case, no analytic model is assumed in advance. Here, the nonlinear and adaptive behavior (and its limitations) strictly emerge from data, and not from a statistically fitted model with a particularly convenient functional form.

An additional issue is answering to the question of what optimality criterion is using the brain in encoding color information at this (low) abstraction level. In this respect, {\em with the considered color image database}, better agreement with the experimental data is obtained using the {\em error minimization} strategy. As expected from its design, the {\em infomax} principle gives rise to steeper nonlinear responses, while the experimental nonlinearities (and the {\em error minimization} solutions) are smoother. Better reproduction of corresponding pairs is also obtained with the {\em error minimization} strategy. These results seem to favor the MacLeod suggestions on \emph{error minimization} in front of the {\em infomax} principle.

A possible objection to such preliminary conclusion would be related to its dependence on the particular dataset.
Note that the steepness of the statistically derived chromatic responses depends on the relative concentration around the achromatic axis in the considered database: if the database would be strongly biased towards achromatic objects, the higher concentration around the achromatic axis would favor the {\em error minimization} strategy. And the other way around for a database of highly saturated objects.
Even though we subscribe the dependence of the results on the database, it does not seem that our database is particularly biased towards achromatic objects (cf. Fig.~\ref{database}). In fact, neglecting the cluster towards saturated green (due to over representation of plants), the curved cluster visible in the CIExy diagram under D65 is quite consistent with the theoretical predictions
in~\cite{Koenderink10}. Therefore, the database does not seem to be specifically favoring the \emph{error minimization} strategy.
Nevertheless, given the practical impossibility of achieving a truly unbiased database~\cite{Koenderink10},
the definitive way to confirm these suggestions on the optimality strategy is extending the Webster and Mollon's measures~\cite{Webster97} performing both color discrimination and corresponding-pairs experiments in which observers are adapted to the same (controlled) statistics as the ones used in the numerical simulations. Recent experiments in color discrimination seem to follow this direction~\cite{Hansen08,Hansen09}. In such experiments, the analysis we proposed here could be used to obtain some insight into the question of the particular optimality criterion applied by the brain in these tasks.


\section{Conclusions}
\label{conclusions}

In this work, we have shown that the basic features of color vision sensors, namely their nonlinearities, the variation of their response under change of adaptation conditions, and their ability to compensate for the changes in spectral illumination emerge from the description of the manifolds of tristimulus values of natural objects under different illuminations. To this end, we have proposed a new nonlinear manifold description technique, the Sequential Principal Curves Analysis with local metric. SPCA is better suited to the color statistics problem than previous manifold description techniques since it is readily invertible and can be easily tuned for the {\em infomax} or the {\em error minimization} principles by simply selecting the appropriate population-dependent metric. In addition, a new accurate color image database has been collected that could be eventually used for further accurate experiments on color constancy and chromatic adaptation.

The proposed technique generalizes previous statistical explanations of color perception that just account for a subset of the data~\cite{Laughlin83,Sejnowski01,Doi03,Atick93,Webster97,Twer01,MacLeod03,MacLeod03b}. Moreover, it also generalizes the results in~\cite{Abrams07}, which simultaneously analyze discrimination and adaptation, because we do not assume an explicit functional form of the model. Consistently with the results of Abrams et al. on the performance of statistically fitted analytic models~\cite{Abrams07}, the proposed non-parametric technique also requires similarity between the reflectance ensembles for an accurate color adaptation.

The simulation of perceptual results with the considered image database suggests that color vision mechanisms may be guided by an {\em error minimization} strategy. However, in order to confirm this conjecture, new psychophysical data are required in which the observers discrimination and adaptation is determined by particular statistics. In that case, SPCA could be applied to obtain some insight about the goal used by the brain in encoding color information.


\end{document}